\documentclass[runningheads]{llncs}
\usepackage{graphicx}
\usepackage{tikz}
\usepackage{comment}
\usepackage{amsmath,amssymb} %
\usepackage{color,soul}
\usepackage{enumitem}
\usepackage{subcaption}
\usepackage{wrapfig}

\usepackage[width=122mm,left=12mm,paperwidth=146mm,height=193mm,top=12mm,paperheight=217mm]{geometry}

\usepackage[breaklinks=true,colorlinks,bookmarks=false]{hyperref}
\usepackage{cite}

\renewcommand{\paragraph}[1]{\noindent {\bf #1}}

\def\preprint{}

\ifdefined\preprint
    \newcommand{\appref}[2]{\ref{#1}}
\else
    \newcommand{\appref}[2]{\textcolor{red}{#2}}
\fi

\begin{document}
\pagestyle{headings}
\mainmatter

\title{Controlling Style and Semantics in Weakly-Supervised Image Generation} %

\titlerunning{Controlling Style and Semantics in Weakly-Supervised Image Generation}
\author{Dario Pavllo \and
Aurelien Lucchi \and
Thomas Hofmann}
\authorrunning{D. Pavllo et al.}
\institute{Department of Computer Science, ETH Zurich}
\maketitle

\begin{center}
    \centering
    \captionsetup{type=figure}
    \includegraphics[width=0.96\textwidth, trim=0 0.25mm 0 1.25mm, clip]{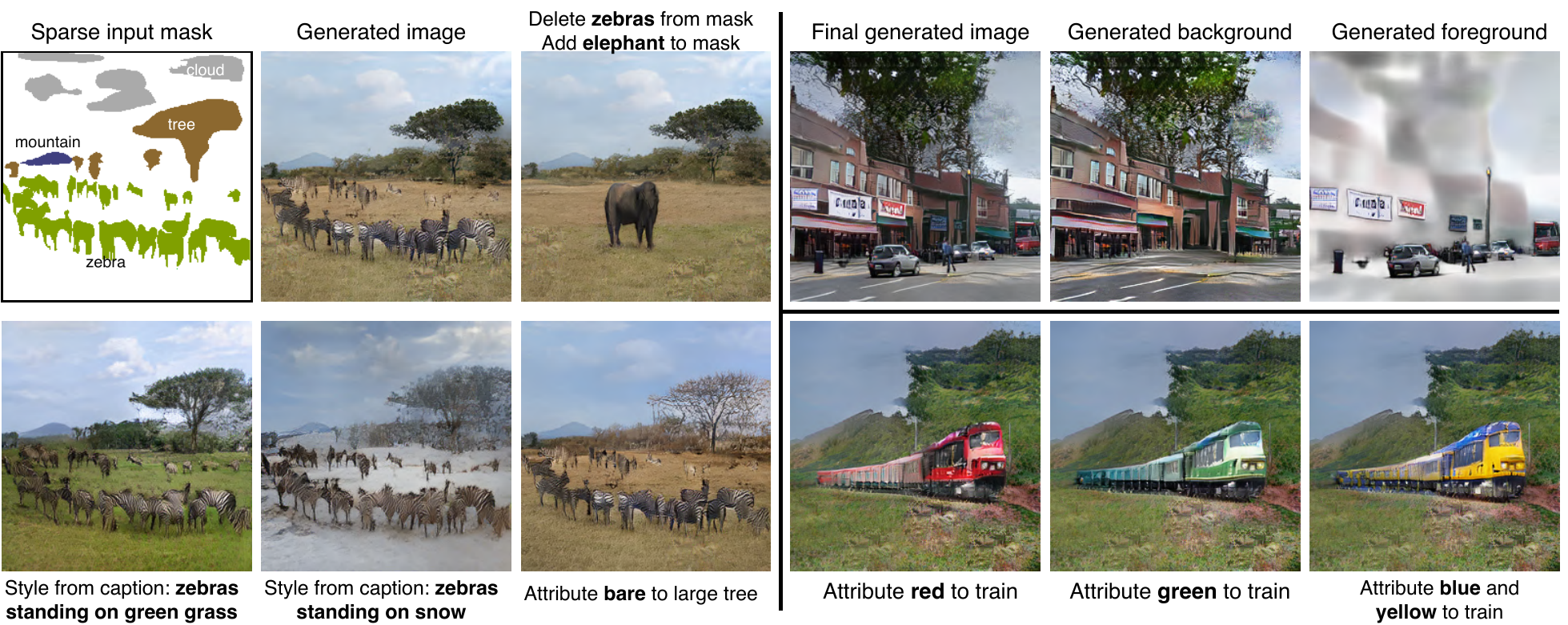}
    \captionof{figure}{\small{Our approach enables control over the style of a scene and its objects via high-level attributes or textual descriptions. It also allows for image manipulation through the mask, including moving, deleting, or adding object instances. The decomposition of the background and foreground (top-right corner) facilitates local changes in a scene.}}
    \label{fig:concept}
\end{center}

\begin{abstract}
We propose a weakly-supervised approach for conditional image generation of complex scenes where a user has fine control over objects appearing in the scene. We exploit sparse semantic maps to control object shapes and classes, as well as textual descriptions or attributes to control both local and global style. In order to condition our model on textual descriptions, we introduce a semantic attention module whose computational cost is independent of the image resolution. To further augment the controllability of the scene, we propose a two-step generation scheme that decomposes background and foreground. The label maps used to train our model are produced by a large-vocabulary object detector, which enables access to unlabeled data and provides structured instance information. In such a setting, we report better FID scores compared to fully-supervised settings where the model is trained on ground-truth semantic maps. We also showcase the ability of our model to manipulate a scene on complex datasets such as COCO and Visual Genome.

\end{abstract}

\section{Introduction}

Deep generative models such as VAEs~\cite{kingma2013vae} and GANs~\cite{goodfellow2014gan} have made it possible to learn complex distributions over various types of data, including images and text. For images, recent technical advances \cite{heusel2017ttur,karras2017progressivegan,miyato2018spectral,miyato2018cgans,zhang2018sagan,brock2018biggan} have enabled GANs to produce realistically-looking images for a large number of classes. However, these models often do not provide high-level control over image characteristics such as appearance, shape, texture, or color, and they fail to accurately model multiple (or compound) objects in a scene, thus limiting their practical applications. 
A related line of research aims at disentangling factors of variation~\cite{karras2019stylegan}. While these approaches can produce images with varied styles by injecting noise at different levels, the style factors are learned without any oversight, leaving the user with a loose handle on the generation process. Furthermore, their applicability has only been demonstrated for single-domain images (e.g. faces, cars, or birds).
Some conditional approaches allow users to control the style of an image using either attributes \cite{yan2016attribute2image,he2019attgan} or natural language \cite{zhang2017stackgan,zhang2018stackganpp,xu2018attngan}, but again, these methods only show compelling results on single-domain datasets.

One key aspect in generative modeling is the amount of required semantic information: i) \emph{weak conditioning} (e.g. a sentence that describes a scene) makes the task underconstrained and harder to learn, potentially resulting in incoherent images on complex datasets. On the other hand, ii) \emph{rich semantic information} (e.g. full segmentation masks) yields the best generative quality, but requires more effort from an artist or annotator. The applications of such richly-conditioned models are numerous, including art, animation, image manipulation, and realistic texturing of video games. Existing works in this category \cite{chen2017crn,qi2018sims,isola2017pix2pix,wang2018pix2pixhd,park2019spade} typically require hand-labeled segmentation masks with per-pixel class annotations. Unfortunately, this is not flexible enough for downstream applications such as image manipulation, where the artist is faced with the burden of modifying the semantic mask coherently. Common transformations such as moving, deleting, or replacing an object require instance information (usually not available) and a strategy for infilling the background. Moreover, these models present little-to-no high-level control over the style of an image and its objects.

Our work combines the merits of both weak conditioning and strong semantic information, by relying on both mask-based generation -- using a variant we call \emph{sparse masks} -- and text-based generation -- which can be used to control the style of the objects contained in the scene as well as its global aspects. \autoref{fig:concept} conceptualizes our idea. Our approach uses a large-vocabulary object detector to obtain annotations, which are then used to train a generative model in a weakly-supervised fashion. The input masks are sparse and retain instance information -- making them easy to manipulate -- and can be inferred from images or videos in-the-wild. We additionally contribute a conditioning scheme for controlling the style of the scene and its instances, either using high-level attributes or natural language with an attention mechanism. Unlike prior approaches, our attention model is applied directly to semantic maps (making it easily interpretable) and its computational cost does not depend on the image resolution, enabling its use in high-resolution settings. This conditioning module is general enough to be plugged into existing architectures.
We also tackle another issue of existing generative models: local changes made to an object (such as moving or deleting) can affect the scene globally due to the learned correlations between classes. While these entangled representations improve scene coherence, they do not allow the user to modify a local part of a scene without affecting the rest. To this end, our approach relies on a multi-step generation process where we first generate a \emph{background image} and then we generate foreground objects conditioned on the former. The background can be frozen while manipulating foreground objects.

Finally, we evaluate our approach on COCO \cite{lin2014mscoco,chen2015mscococaptions,caesar2018cocostuff} and Visual Genome \cite{krishna2017visualgenome}, and show that our weakly-supervised setting can achieve better FID scores \cite{heusel2017ttur} than fully-supervised counterparts trained on ground-truth masks, and weakly-supervised counterparts where the model is trained on dense maps obtained from an off-the-shelf semantic segmentation model, while being more controllable and scalable to large unlabeled datasets. We show that this holds both in presence and in absence of style control.\\
Code is available at \url{https://github.com/dariopavllo/style-semantics}.

\section{Related work}
The recent success of GANs has triggered interest for conditional image synthesis from categorical labels~\cite{miyato2018spectral,miyato2018cgans,zhang2018sagan,brock2018biggan}, text~\cite{reed2016generative,zhang2017stackgan,zhang2018stackganpp,xu2018attngan}, semantic maps~\cite{isola2017pix2pix,wang2018pix2pixhd,park2019spade}, and conditioning images from other domains~\cite{zhu2017cyclegan,isola2017pix2pix}.

\paragraph{Image generation from semantic maps.} In this setting, a semantic segmentation map is translated into a natural image. Non-adversarial approaches are typically based on perceptual losses \cite{chen2017crn,qi2018sims}, whereas GAN architectures are based on patch-based discriminators \cite{isola2017pix2pix}, progressive growing \cite{wang2018pix2pixhd,karras2017progressivegan}, and conditional batch normalization where the semantic map is fed to the model at different resolutions \cite{park2019spade}. Similarly to other state-of-the-art methods, our work is also based on this paradigm. Most approaches are trained on hand-labeled masks (limiting their application in the wild), but \cite{park2019spade} shows one example where the model is weakly supervised on masks inferred using a semantic segmentation model \cite{chen2017deeplab}. Our model is also weakly supervised, but instead of a semantic segmentation model we use an object detector -- which allows us to maintain instance information during manipulations, and results in \emph{sparse masks}. While early work focused on class semantics, recent methods support some degree of style control. E.g.\ \cite{wang2018pix2pixhd} trains an instance autoencoder and allows the user to choose a latent code from among a set of modes, whereas \cite{park2019spade} trains a VAE to control the global style of a generated image by copying the style of a \emph{guide} image. Both these methods, however, do not provide fine-grained style control (e.g. changing the color of an object to \emph{red}).
Another recent trend consists in generating images from structured layouts, which are transformed into semantic maps as an intermediate step to facilitate the task. In this regard, there is work on generation from bounding-box layouts \cite{hong2018inferring,zhao2019imagelayout,hinz2019generatingdistinct,sun2019image} and scene graphs \cite{johnson2018imagescenegraphs}. Although these approaches tackle a harder task, they generate low-resolution images and are not directly relatable to our work, which tackles controllability among other aspects.

\begin{figure}[t]
	\centering
    \includegraphics[width=\linewidth, trim=0 1mm 0 1.5mm, clip]{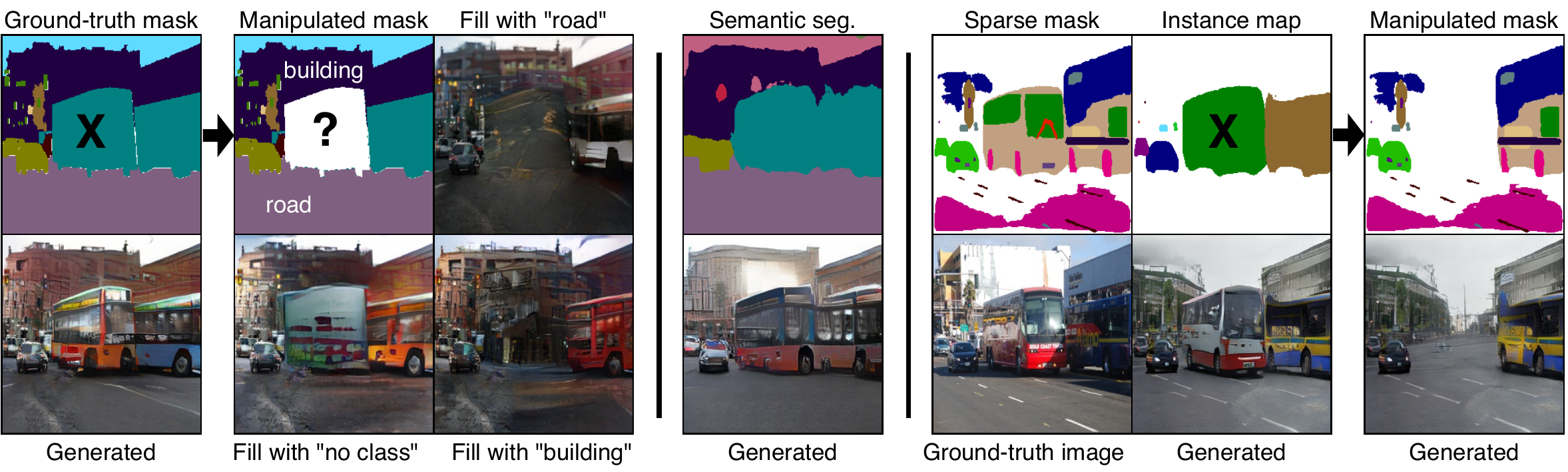}
    \caption{\small \textbf{Left:} when manipulating a ground-truth mask (e.g. deleting one bus), one is left with the problem of infilling the background which is prone to ambiguities (e.g. selecting a new class as either road or building). Furthermore, in existing models, local changes affect the scene globally due to learned correlations. \textbf{Middle:} in the wild, ground-truth masks are not available (neither are instance maps). One can infer maps using a semantic segmentation model, but these are often noisy and lack instance information (in the example above, we observe that the two buses are merged). \textbf{Right:} our weakly-supervised sparse mask setting, which combines fine-detailed masks with instance information. The two-step decomposition ensures that changes are localized.}
	\label{fig:manipulations}
\end{figure}

\paragraph{Semantic control.}
Existing approaches do not allow for easy manipulation of the semantic map because they present no interface for encoding existing images. In principle, it is possible to train a weakly-supervised model on maps inferred from a semantic segmentation model, as \cite{park2019spade} does for landscapes. However, as we show in \autoref{sec:results}, the results in this setting are notably worse than fully-supervised baselines. Furthermore, manipulations are still challenging because instance information is not available. Since the label masks are \emph{dense}, even simple transformations such as deleting or moving an object would create holes in the semantic map that need to be adjusted by the artist (\autoref{fig:manipulations}). \emph{Dense} masks also make the task too constrained with respect to background aspects of the scene (e.g. sky, land, weather), which leaves less room for style control. Semantic control can also be framed as an unpaired image-to-image translation task \cite{mo2018instagan}, but this requires ground-truth masks for both source and target instances, and can only translate between two classes.

\paragraph{Text-based generation.}
Some recent models condition the generative process on text data. These are often based on autoregressive architectures~\cite{reed2016generating} and GANs~\cite{reed2016generative,zhang2017stackgan,zhang2018stackganpp,xu2018attngan}.
Learning to generate images from text using GANs is known to be difficult due to the task being unconstrained. In order to ease the training process, \cite{zhang2017stackgan,zhang2018stackganpp} propose a two-stage architecture named \emph{StackGAN}. To avoid the instability associated with training a language model jointly with a GAN, they use a pretrained sentence encoder \cite{kiros2015skipthought} that encodes a caption into a fixed-length vector which is then fed to the model. More advanced architectures such as \emph{AttnGAN} \cite{xu2018attngan} use an attention mechanism which we discuss in one of the next paragraphs. These approaches show interesting results on single-domain datasets (birds, flowers, etc.) but are less effective on complex datasets such as \emph{COCO} \cite{lin2014mscoco} due to the intrinsic difficulty of generating coherent scenes from text alone.
Some works~\cite{joseph2019c4synth,yin2019semanticsdisentangling} have demonstrated that generative models can benefit from taking as input multiple diverse textual descriptions per image. Finally, we are not aware of any prior work that conditions the generative process on \emph{both} text and semantic maps (our setting).

\paragraph{Multi-step generation.} %
Approaches such as \cite{yang2017lrgan,singh2018finegan} aim at disentangling background and foreground generation. While fully-unsupervised disentanglement is provably impossible \cite{locatello2018challenging}, it is still achievable through some form of inductive bias -- either in the model architecture or in the loss function. While \cite{yang2017lrgan} uses spatial transformers to achieve separation, \cite{singh2018finegan} uses object bounding boxes. Both methods show compelling results on single-domain datasets that depict a centered object, but are not directly applicable to more challenging datasets. For composite scenes, \cite{turkoglu2019layer} generates foreground objects sequentially to counteract merging effects. In our work, we are not interested in full disentanglement (i.e. we do not assume independence between background and foreground), but merely in separating the two steps while keeping them interpretable. Our model still exploits correlations among classes to maximize visual quality, and is applied to datasets with complex scenes.
Finally, there has also been work on interactive generation using dialogue \cite{elnouby2019geneva,cheng2018sequentialgandialogue,sharma2018chatpainter}.

\paragraph{Attention models in GANs.} For unconditional models (or models conditioned on simple class labels), self-attention GANs \cite{zhang2018sagan,brock2018biggan} use visual-visual attention to improve spatial coherence. For generation from text, \cite{xu2018attngan} employ sentence-visual attention coupled with an LSTM  encoder, but only in the generator. In the discriminator, the caption is enforced through a supervised loss based on features extracted from a pretrained Inception \cite{szegedy2016inception} network. We introduce a new form of attention (\emph{sentence-semantic}) which is applied to semantic maps instead of convolutional feature maps, and whose computational cost is independent of the image resolution. It is applied both to the generator and the discriminator, and on the sentence side it features a transformer-based \cite{vaswani2017transformer} encoder.

\section{Approach}
\subsection{Framework}
\label{sec:framework}
Our main interest is conditional image generation of complex scenes where a user has fine control over the objects appearing in the scene. Prior work has focused on generating objects from ground-truth masks \cite{zhu2017cyclegan,isola2017pix2pix,wang2018pix2pixhd,park2019spade} or on generating outdoor scenes based on simple hand-drawn masks \cite{park2019spade}. While the former approach requires a significant labeling effort, the latter is not directly suitable for complex datasets such as COCO-Stuff \cite{caesar2018cocostuff}, whose images consist of a large number of classes with complex (hard to draw) shapes.
We address these problems by introducing a new model that is conditioned on sparse masks -- to control object shapes and classes -- and on text/attributes to control style and textures. This gives the ability to a user to produce scenes through a variety of image manipulations (such as moving, scaling or deleting an instance, adding an instance from another image or from a database of shapes) as well as style manipulations controlled using either high-level attributes on individual instances (e.g. \emph{red}, \emph{green}, \emph{wet}, \emph{shiny}) or using text that refers to objects as well as global context (e.g. ``a red car at night''). In the latter case, visual-textual correlations are not explicitly defined but are learned in an unsupervised way.

\paragraph{Sparse masks.}
Instead of training a model on precise segmentation masks as in \cite{isola2017pix2pix,wang2018pix2pixhd,park2019spade}, we use a mask generated automatically from a large-vocabulary object detector. Compared to a weakly-supervised setting based on semantic segmentation, this process introduces less artifacts (see Appendix \appref{sec:appendix-results}{A.4} in the supplementary material) and has the benefit of providing information about each instance (which may not always be available otherwise), including parts of objects which would require significant manual effort to label in a new dataset.
In general, our set of classes comprises countable objects (person, car, etc.), parts of objects (light, window, door, etc.), as well as uncountable classes (grass, water, snow), which are typically referred to as ``stuff'' in the COCO terminology \cite{caesar2018cocostuff}. For the latter category, an object detector can still provide useful sparse information about the background, while keeping the model autonomous to fill-in the gaps. We describe the details of our object detection setup in \autoref{sec:implementation-details}.

\begin{figure*}[b]
    \centering
    \includegraphics[width=\textwidth]{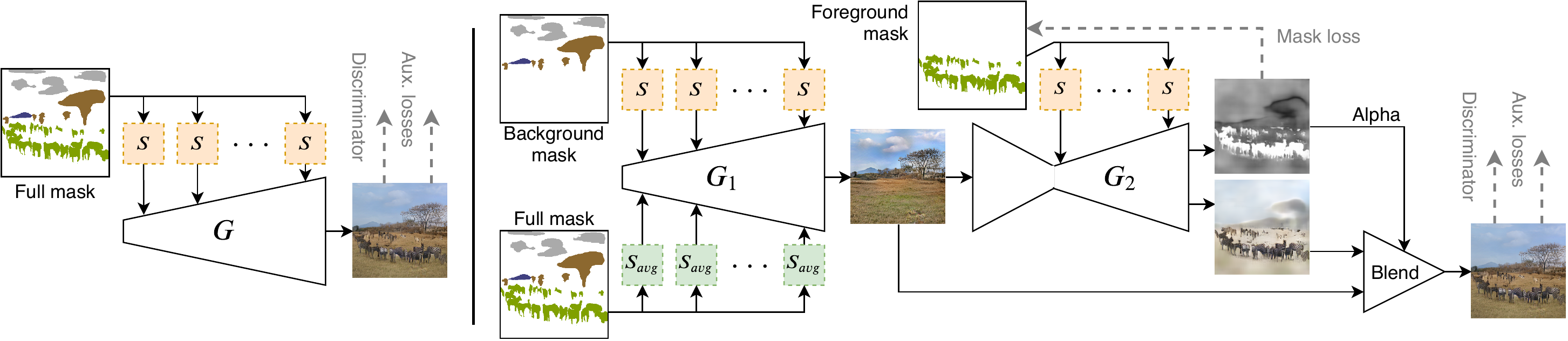}
	\caption{\small \textbf{Left:} One-step model. \textbf{Right:} two-step model. The background generator $G_1$ takes as input a \emph{background mask} (processed by $S$-blocks) and the full mask (processed by $S_{avg}$-blocks, where positional information is removed). The foreground generator takes as input the output of $G_1$ and a \emph{foreground mask}. Finally, the two outputs are alpha-blended. For convenience, we do not show attributes/text in this figure.}
	\label{fig:architectures}
\end{figure*}

\paragraph{Two-step generation.} In the absence of constraints, conditional models learn class correlations observed in the training data. For instance, while dogs typically stand on green grass, zebras stand on yellow grass. While this feature is useful for maximizing scene coherence, it is undesirable when only a local change in the image is wanted. We observed similar global effects on other local transformations, such as moving an object or changing its attributes, and generally speaking, small perturbations of the input can result in large variations of the output. We show a few examples in the Appendix \appref{sec:appendix-results}{A.4}.
To tackle this issue, we propose a variant of our architecture which we call \emph{two-step} model and which consists of two concatenated generators (\autoref{fig:architectures}, right). The first step (generator $G_1$) is responsible for generating a \emph{background} image, whereas the second step (generator $G_2$) generates a \emph{foreground} image conditioned on the background image. The definition of what constitutes background and foreground is arbitrary: our choice is to separate by class: static/uncountable objects (e.g. buildings, roads, grass, and other surfaces) are assigned to \emph{background}, and moving/countable objects are assigned to \emph{foreground}. Some classes can switch roles depending on the parent class, e.g. \emph{window} is \emph{background} by default, but it becomes \emph{foreground} if it is a child of a foreground object such as a car.\\
When applying a local transformation to a foreground object, the background can conveniently be frozen to avoid global changes. As a side benefit, this also results in a lower computational cost to regenerate an image.
Unlike work on disentanglement \cite{yang2017lrgan,singh2018finegan} which enforces that the background is independent of the foreground without necessarily optimizing for visual quality, our goal is to enforce separation while maximizing qualitative results. In our setting, $G_1$ is exposed to both background and foreground objects, but its architecture is designed in a way that foreground information is not rendered, but only used to induce a bias in the background (see \autoref{sec:architecture}).

\paragraph{Attributes.} Our method allows the user to control the style of individual instances using high-level attributes. These attributes refer to appearance factors such as colors (e.g. white, black, red), materials (wood, glass), and even modifiers that are specific to classes (leafless, snowy), but not shape or size, since these two are determined by the mask. An object can also combine multiple attributes (e.g. black and white) or have none -- in this case, the generator would pick a predefined mode. This setup gives the user a lot of flexibility to manipulate a scene, since the attributes need not be specified for every object.

\paragraph{Captions.} Alternatively, one can consider conditioning style using natural language. This has the benefit of being more expressive, and allows the user to control global aspects of the scene (e.g. time of the day, weather, landscape) in addition to instance-specific aspects. While this kind of conditioning is harder to learn than plain attributes, in \autoref{sec:architecture} we introduce a new attention model that shows compelling results without excessively increasing the model complexity.

\begin{figure}[t]
    \centering
    \includegraphics[width=0.45\linewidth]{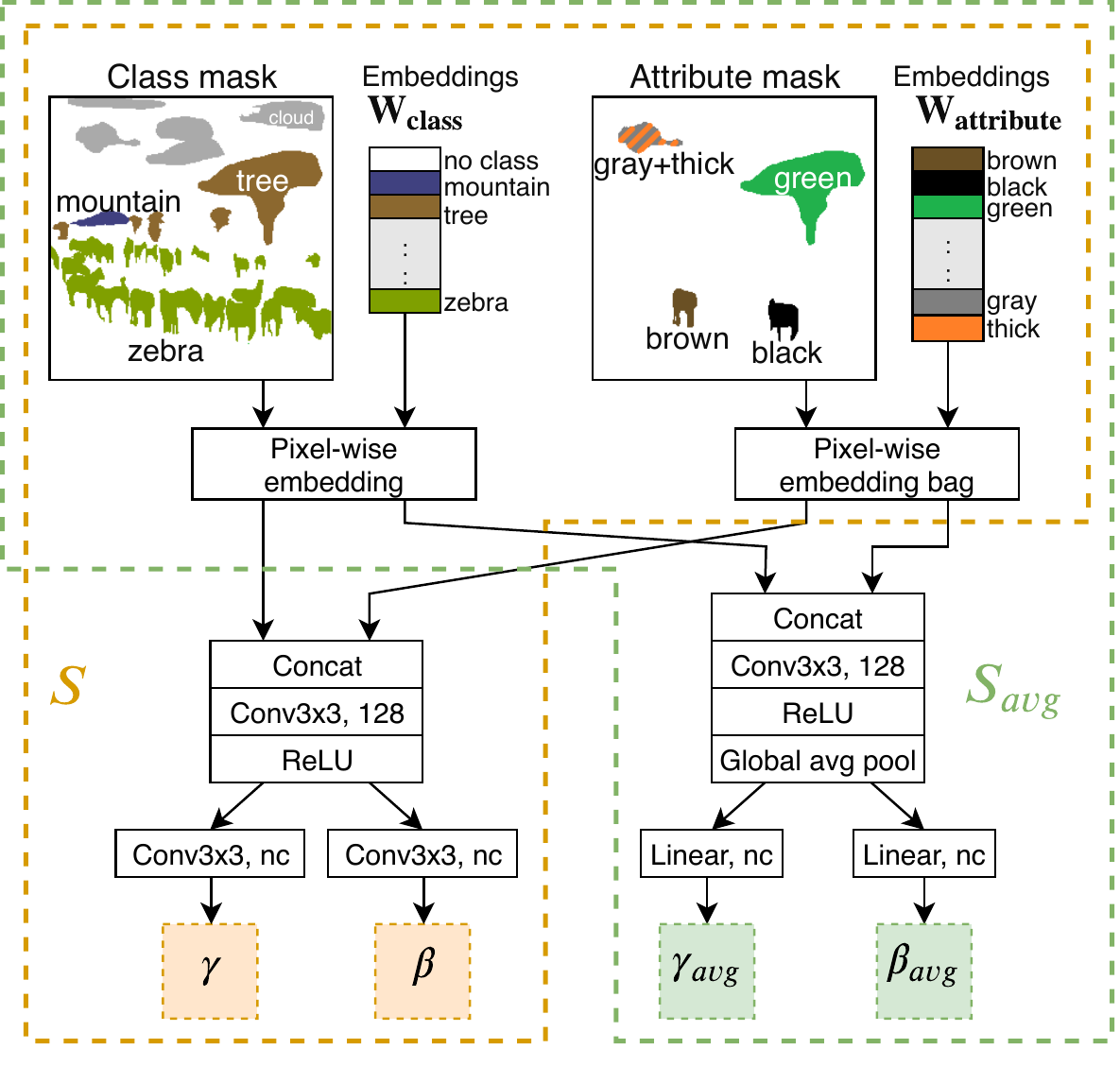}
    \includegraphics[width=0.54\linewidth]{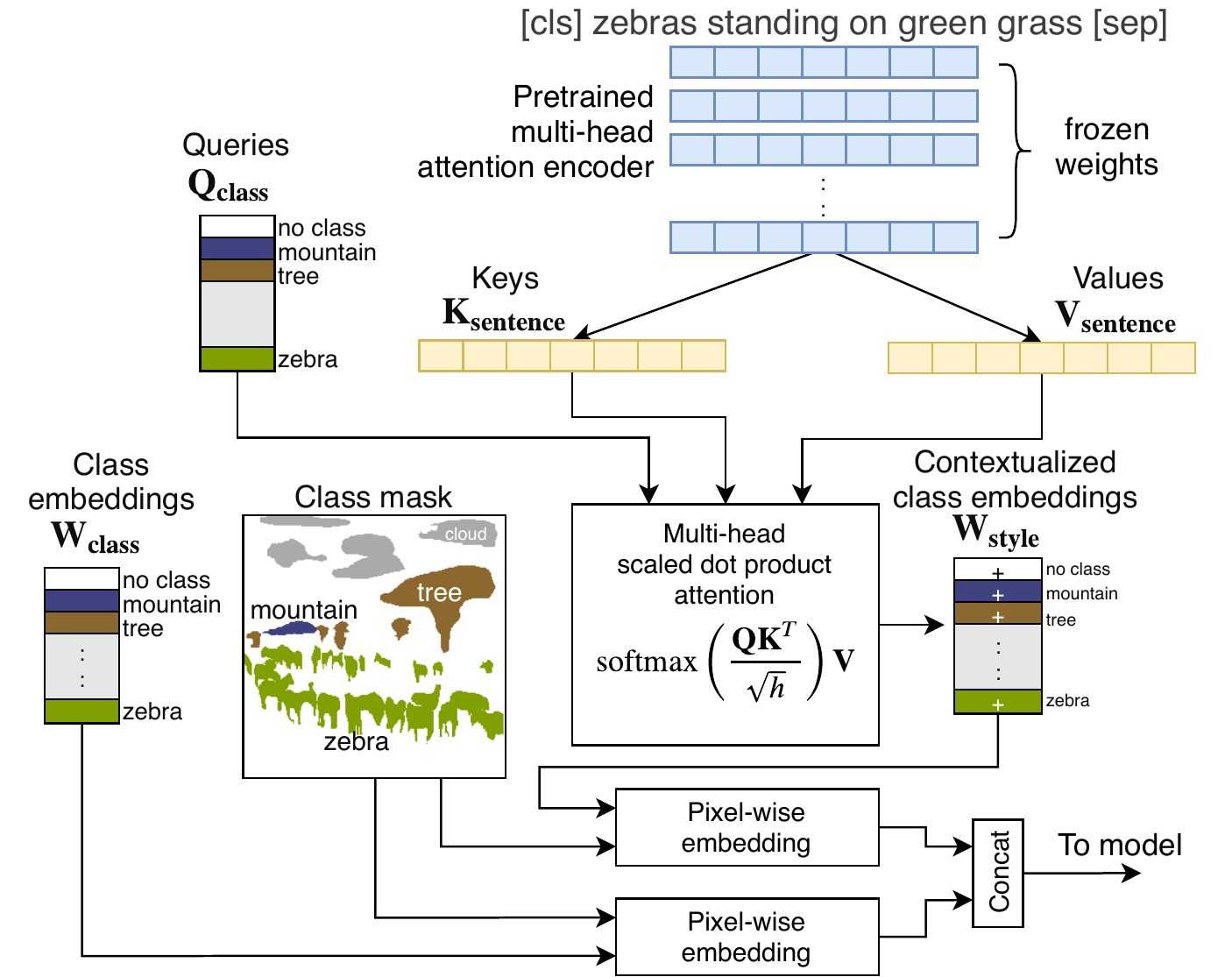}
	\caption{\small \textbf{Left:} Conditioning block with attributes. Class and attribute embeddings are concatenated and processed to generate the conditional batch normalization gain and bias. In the attribute mask, embeddings take the contour of the instance to which they refer. In $G_1$ of the two-step model, where $S$ and $S_{avg}$ are both used, the embedding weights are shared. \textbf{Right:} Attention mechanism for conditioning style via text. The sentence (of length $n = 7$ including delimiters) is fed to a pretrained attention encoder, and each token is transformed into a key and a value using two trainable linear layers. The queries are learned for each class, and the attention yields a set of contextualized class embeddings that are concatenated to the regular semantic embeddings.}
	\label{fig:blocks}
\end{figure}

\subsection{Architecture.}
\label{sec:architecture}
We design our conditioning mechanisms to have sufficient generality to be attached to existing conditional generative models. In our experiments, we choose \emph{SPADE} \cite{park2019spade} as the backbone for our conditioning modules, which to our knowledge represents the state of the art. As in \cite{park2019spade}, we use a multi-scale discriminator \cite{wang2018pix2pixhd}, a perceptual loss in the generator using a pretrained VGG network \cite{simonyan2014vgg}, and a feature matching loss in the discriminator \cite{wang2018pix2pixhd}.

\paragraph{One-step model.} 
Since this model (\autoref{fig:architectures}, left) serves as a baseline, we keep its backbone as close as possible to the reference model of~\cite{park2019spade}. 
We propose to insert the required information about attributes/captions in this architecture by modifying the input layer and the conditional batch normalization layers of the generator, which is where semantic information is fed to the model. We name these \emph{S}-blocks (short for \emph{semantic-style} block).

\paragraph{Semantic-style block.} For class semantics, the input sparse mark is fed to a pixel-wise embedding layer to convert categorical labels into 64D embeddings (including the empty space, which is a special class ``no class''). To add style information, we optionally concatenate another 64D representation to the class embedding (pixel-wise); we explain how we derive this representation in the next two paragraphs. The resulting feature map is convolved with a $3 \times 3$ kernel, passed through a ReLU non-linearity and convolved again to produce two feature maps $\boldsymbol{\gamma}$ and $\boldsymbol{\beta}$, respectively, the conditional batch normalization gain and bias. The normalization is then computed as $\mathbf{y} = \text{BN}(\mathbf{x}) \odot (1 + \boldsymbol{\gamma}) + \boldsymbol{\beta}$, where $\text{BN}(\mathbf{x})$ is the parameter-free batch normalization. The last step is related to \cite{park2019spade} and other architectures based on conditional batch normalization. Unlike \cite{park2019spade}, however, we do not use $3 \times 3$ convolutions on one-hot representations in the input layer. This allows us to scale to a larger number of classes without significantly increasing the number of parameters. We apply the same principle to the discriminators.

\paragraph{Conditioning on attributes.}
For attributes, we adopt a \emph{bag-of-embeddings} approach where we learn a 64D embedding for each possible attribute, and all attribute embeddings assigned to an instance are broadcast to the contour of the instance, summed together, and concatenated to the class embedding. \autoref{fig:blocks} (left) ($S$-block) depicts this process. To implement this efficiently, we create a multi-hot \emph{attribute mask} (1 in the locations corresponding to the attributes assigned to the instance, 0 elsewhere) and feed it through a $1 \times 1$ convolutional layer with $N_{attr}$ input channels and 64 output channels.
Attribute embeddings are shared among classes and are not class-specific. This helps the model generalize better (e.g. colors such as ``white'' apply both to vehicles and animals), and we empirically observe that implausible combinations (e.g. leafless person) are simply ignored by the generator without side effects.

\paragraph{Conditioning on text.}
While previous work has used fixed-length vector representations \cite{zhang2017stackgan,zhang2018stackganpp} or one-layer attention models coupled with RNNs \cite{xu2018attngan}, the diversity of our scenes led us to use a more powerful encoder entirely based on self-attention \cite{vaswani2017transformer}. We encode the image caption using a pretrained BERT$_{base}$ model \cite{devlin2018bert} (110M parameters). It is unreasonable to attach such a model to a GAN and fine-tune it, both due to excessive memory requirements and due to potential instabilities. Instead, we freeze the pretrained model and encode the sentence, extract its hidden representation after the last or second-to-last layer (we compare these in \autoref{sec:results}), and train a custom multi-head attention layer for our task. This paradigm, which is also suggested by \cite{devlin2018bert}, has proven successful on a variety of NLP downstream tasks, especially when these involve small datasets or limited vocabularies. Furthermore, instead of storing the language model in memory, we simply pre-compute the sentence representations and cache them.

Next, we describe the design of our trainable attention layer (\autoref{fig:blocks}, right). Our attention mechanism is different from the commonly-used sentence-visual attention \cite{xu2018attngan}, where attention is directly applied to convolutional feature maps inside the generator. Instead, we propose a form of sentence-semantic attention which is computationally efficient, interpretable, and modular. It can be concatenated to conditioning layers in the same way as we concatenate attributes. Compared to sentence-visual attention, whose cost is $\mathcal{O}(nd^2)$ (where $n$ is the sentence length and $d \times d$ is the feature map resolution), our method has a cost of $\mathcal{O}(nc)$ (where $c$ is the number of classes), i.e. it is independent of the image resolution.
We construct a set of $c$ \emph{queries} (i.e. one for each class) of size $h = 64$ (where $h$ is the attention head size). We feed the hidden representations of each token of the sentence to two linear layers, one for the \emph{keys} and one for the \emph{values}. Finally, we compute a scaled dot-product attention \cite{vaswani2017transformer}, which yields a set of $c$ \emph{values}. To allow the conditioning block to attend to multiple parts of the sentence, we use 6 or 12 attention heads (ablations in \autoref{sec:results}), whose output values are concatenated and further transformed through a linear layer. This process can be thought of as generating \emph{contextualized class embeddings}, i.e. class embeddings customized according to the sentence. For instance, given a semantic map that depicts a car and the caption ``a red car and a person'', the query corresponding to the visual class \emph{car} would most likely attend to ``red car'', and the corresponding value will induce a bias in the model to add redness to the position of the car. Finally, the \emph{contextualized class embeddings} are applied to the semantic mask via pixel-wise matrix multiplication with one-hot vectors, and concatenated to the \emph{class embeddings} in the same way as attributes. In the current formulation, this approach is unable to differentiate between instances of the same class. We propose a possible mitigation in \autoref{sec:conclusion}.

\paragraph{Two-step model.} It consists of two concatenated generators. $G_1$ generates the background, i.e. it models $p(x_\text{bg})$, whereas $G_2$ generates the foreground conditioned on the background, i.e. $p(x_\text{fg} | x_\text{bg})$. One notable difficulty in training such a model is that background images are never observed in the training set (we only observe the final image), therefore we cannot use an intermediate discriminator for $G_1$. Instead, we use a single, final discriminator and design the architecture in a way that the gradient of the discriminator (plus auxiliary losses) is redirected to the correct generator. The convolutional nature of $G_1$ would then ensure that the background image does not contain visible holes. A natural choice is \emph{alpha blending}, which is also used in \cite{yang2017lrgan,singh2018finegan}. $G_2$ generates an RGB foreground image plus a transparency mask (\emph{alpha} channel), and the final image is obtained by pasting the foreground onto the background via linear blending:
\begin{align}
\label{eq:alphablend}
    x_{\text{final}} = x_{\text{bg}} \cdot (1 - \alpha_{\text{fg}}) + x_{\text{fg}} \cdot \alpha_{\text{fg}}
\end{align}
where $x_{\text{final}}$, $x_{\text{bg}}$, and $x_{\text{fg}}$ are RGB images, and $\alpha_{\text{fg}}$ is a 1-channel image bounded in $[0, 1]$ by a sigmoid.
Readers familiar with highway networks \cite{srivastava2015highway} might notice a similarity to this approach in terms of gradients dynamics. If $\alpha_{\text{fg}} = 1$, the gradient is completely redirected to $x_{\text{fg}}$, while if $\alpha_{\text{fg}} = 0$, the gradient is redirected to $x_{\text{bg}}$. This scheme allows us to train both generators in an end-to-end fashion using a single discriminator, and we can also preserve auxiliary losses (e.g.\ VGG loss) which \cite{park2019spade} has shown to be very important for convergence. To incentivize separation between classes as defined in \autoref{sec:framework}, we supervise $\alpha_{fg}$ using a binary cross-entropy loss, and decay this term over time (see \autoref{sec:implementation-details}).

$G_2$ uses the same \emph{S}-blocks as the ones in the one-step model, but here they take a \emph{foreground mask} as input (\autoref{fig:architectures}, right). $G_1$, on the other hand, must exploit foreground information without rendering it. We therefore devise a further variation of input conditioning that consists of two branches: (i) the first branch ($S$-block) takes a \emph{background mask} as input and processes it as usual to produce the batch normalization gain $\boldsymbol{\gamma}$ and bias $\boldsymbol{\beta}$. (ii) the second branch ($S_{avg}$-block, \autoref{fig:blocks} left) takes the full mask as input (background plus foreground), processes it, and applies global average pooling to the feature map to remove information about localization. This way, foreground information is only used to bias $G_1$ and cannot be rendered at precise spatial locations. After pooling, it outputs $\boldsymbol{\gamma_{avg}}$ and $\boldsymbol{\beta_{avg}}$. (iii) The final conditional batch normalization is computed as:
\begin{equation}
    \mathbf{y} = \text{BN}(\mathbf{x}) \odot (1 + \boldsymbol{\gamma} + \boldsymbol{\gamma_{avg}}) + \boldsymbol{\beta} + \boldsymbol{\beta_{avg}}
\end{equation}
Finally, the discriminator $D$ takes the full mask as input (background plus foreground). Note that, if $G_1$ took the full mask as input without information reduction, it would render visible ``holes'' in the output image due to gradients never reaching the foreground zones of the mask, which is what we are trying to avoid.  The Appendix \appref{sec:appendix-architecture}{A.1} provides more details about our architectures, and \appref{sec:appendix-inference}{A.2} shows how $G_2$ can be used to generate one object at a time to fully disentangle foreground objects from each other (although this is unnecessary in practice). %

\section{Experiments}
\label{sec:experiments}
For consistency with~\cite{park2019spade}, we always evaluate our model on the COCO-Stuff validation set \cite{caesar2018cocostuff}, but we train on a variety of training sets:

\noindent\textbf{COCO-Stuff (COCO2017)} \cite{lin2014mscoco,caesar2018cocostuff} contains 118k training images with captions \cite{chen2015mscococaptions}. We train with and without captions. COCO-Stuff extends COCO2017 with ground-truth semantic maps, but for our purposes the two datasets are equivalent since we do not exploit ground-truth masks.\\
\textbf{Visual Genome (VG)} \cite{krishna2017visualgenome} contains 108k images that partially overlap with COCO ($\approx$50\%). VG does not have a standard train/test split, therefore we leave out 10\% of the dataset to use as a validation set (IDs ending with 9), and use the rest as a training set from which we remove images that overlap with the COCO-Stuff validation set. We extract the attributes from the scene graphs.\\
\textbf{Visual Genome augmented (VG+)} VG augmented with the 123k images from the COCO unlabeled set. The total size is 217k images after removing exact duplicates. The goal is to evaluate how well our method scales to large unlabeled datasets. We train without attributes and without captions.

For all experiments, we evaluate the \emph{Fréchet Inception Distance} (FID) \cite{heusel2017ttur} (precise implementation details of the FID in the Appendix \appref{sec:appendix-fid}{A.3}). Furthermore, we report our results in \autoref{sec:results} and provide additional qualitative results in \appref{sec:appendix-results}{A.4}.

\subsection{Implementation details}
\label{sec:implementation-details}
\paragraph{Semantic maps.}
To construct the input semantic maps, we use the semi-supervised implementation of Mask R-CNN \cite{he2017maskrcnn,ren2015fasterrcnn} proposed by \cite{hu2018segmenteverything}. It is trained on bounding boxes from Visual Genome (3000 classes) and segmentation masks from COCO (80 classes), and learns to segment classes for which there are no ground-truth masks. We discard the least frequent classes, and, since some VG concepts overlap (e.g. car, vehicle) leading to spurious detections, we merge these classes and end up with a total of $c = 280$ classes (plus a special class for ``no class''). We set the threshold of the object detector to 0.2, and further refine the predictions by running a class-agnostic non-maximum-suppression (NMS) step on the detections whose mask intersection-over-union (IoU) is greater than 0.7. We also construct a transformation hierarchy to link children to their parents in the semantic map (e.g. headlight of a car) so that they can be manipulated as a whole; further details in the Appendix~\appref{sec:appendix-architecture}{A.1}.
We select the 256 most frequent attributes, manually excluding those that refer to shapes (e.g. \emph{short}, \emph{square}).

\begin{figure*}[t]
    \centering
    \includegraphics[width=\textwidth, trim=0 0 0 2mm, clip]{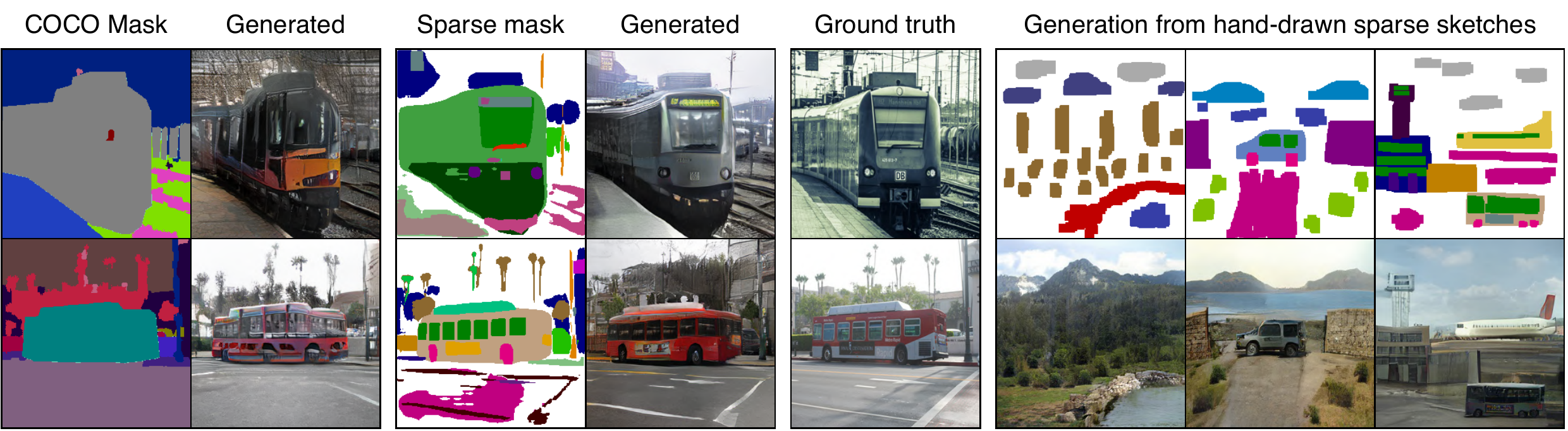}
	\caption{\small \textbf{Left:} the larger set of labels in our sparse masks improves fine details. These masks are easy to obtain with a semi-supervised object detector, and would otherwise be too hard to hand-label. \textbf{Right:} sparse masks are also easy to sketch by hand.}
	\label{fig:coco-vs-sparse}
\end{figure*}

\paragraph{Training.} We generate images at 256$\times$256 and keep our experimental setting and hyperparameters as close as possible to \cite{park2019spade} for a fair comparison. For the two-step model, we provide supervision on the alpha blending mask and decay this loss term over time, observing that the model does not re-entangle background and foreground. This gives $G_2$ some extra flexibility in drawing details that are not represented by the mask (reflections, shadows). Hyperparameters and additional training details are specified in the Appendix \appref{sec:appendix-architecture}{A.1}.

\begin{figure*}[t]
    \centering
    \includegraphics[width=\textwidth, trim=0 0.5mm 0 1.5mm, clip]{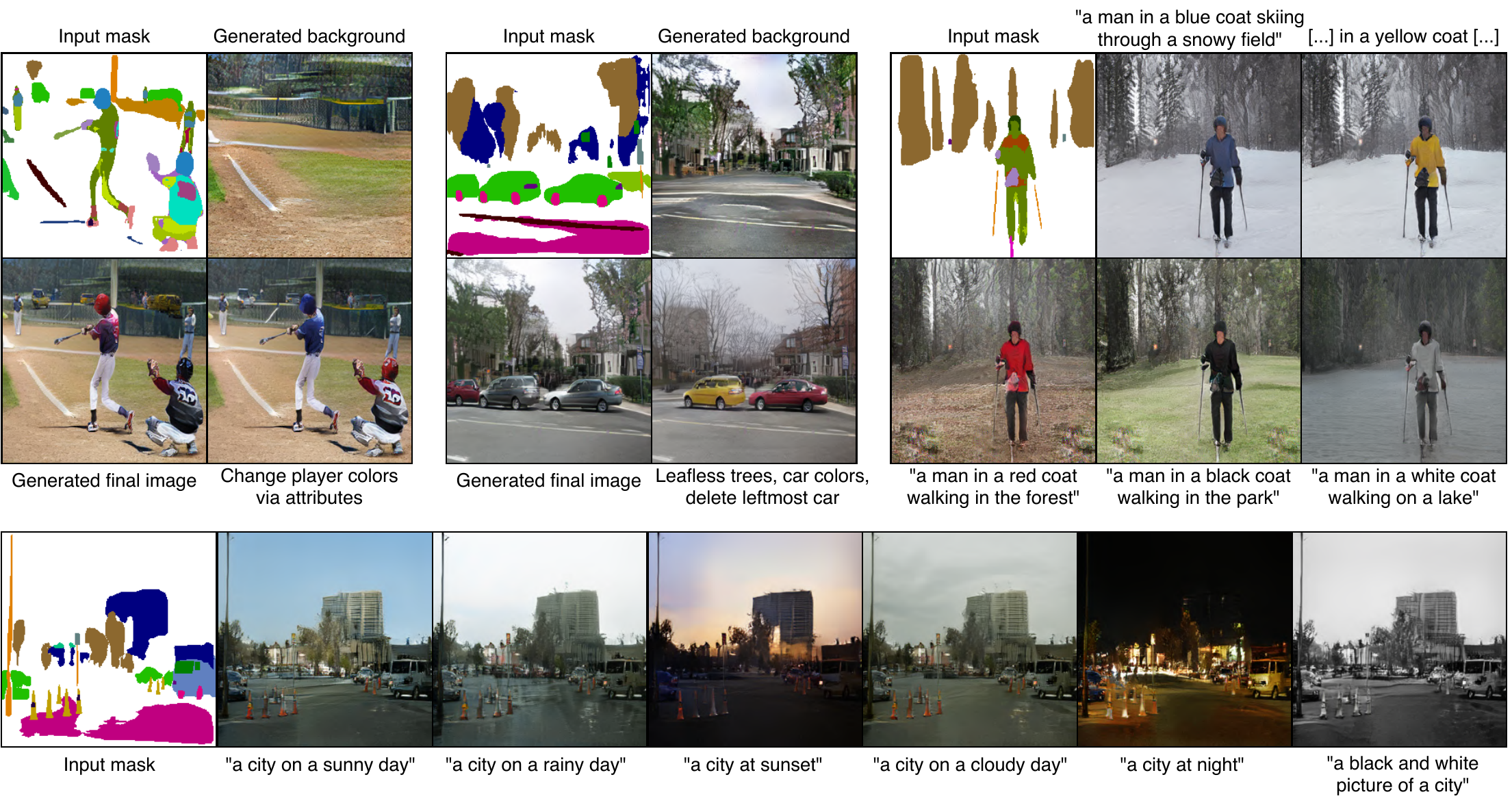}
	\caption{\small Qualitative results ($256 \times 256$). \textbf{Top-left} and \textbf{top-middle}: two-step generation with manipulation of attributes and instances. \textbf{Top-right:} manipulating style (both context and instances) via text. \textbf{Bottom:} manipulating global style via text.}
	\label{fig:qualitative-results}
\end{figure*}

\subsection{Results}
\label{sec:results}
\paragraph{Quantitative.} We show the FID scores for the main experiments in \autoref{tab:fid-scores} (left). While improving FID scores is not the goal of our work, our weakly-supervised sparse mask baseline (\#3) interestingly outperforms both the fully-supervised baseline on SPADE \cite{park2019spade} (\#1) and the weakly-supervised baseline (\#2) trained on dense semantic maps. These experiments adopt an identical architecture and training set, no style input, and differ only in the type of input mask. For \#2 we obtain the semantic maps from \emph{DeepLab-v2} \cite{chen2017deeplab}, a state-of-the-art semantic segmentation model pretrained on COCO-Stuff. Our improvement is partly due to masks better representing fine details (such as windows, doors, lights, wheels) in compound objects, which are not part of the COCO class set. In \autoref{fig:coco-vs-sparse} (left) we show some examples. Moreover, the experiment on the \emph{augmented} Visual Genome dataset highlights that our model benefits from extra unlabeled images (\#4).
Rows \#5--9 are trained with style input. In particular, we observe that these outperform the baseline even when they use a two-step architecture (which is more constrained) or are trained on a different training set (VG instead of COCO). Row \#6-7 draw their text embeddings from the last BERT layer and adopt 12 attention heads (the default), whereas \#5 draws its embeddings from the 2nd-last layer, uses 6 heads, and performs slightly better.

\paragraph{Qualitative.} In \autoref{fig:qualitative-results} we show qualitative results as well as examples of manipulations, either through attributes or text. Additional examples can be seen in the Appendix \appref{sec:appendix-results}{A.4}, including latent space interpolation \cite{kilcher2018semantic}. In \appref{sec:appendix-attention}{A.5}, we visualize the attention mechanism. Finally, we observe that sketching sparse masks by hand is very practical (\autoref{fig:coco-vs-sparse}, right) and provides an easier interface than dense semantic maps (in which the class of every pixel must be manually specified). The supplementary video (see Appendix \appref{sec:appendix-video}{A.7}) shows how these figures are drawn.

\begin{table}[t]
    \centering
	\caption{\small \textbf{Left:} FID scores for the main experiments; lower is better. The first line represents the SPADE baseline \cite{park2019spade}. For the models trained on VG, we also report FID scores on our VG validation set. ($\dagger$) indicates that the model is weakly-supervised, (6h) denotes ``6 attention heads'', $L_{n-1}$ indicates that the text embeddings are drawn from the second-to-last BERT layer. \textbf{Right:} ablation study with extra experiments.}
	\vspace{1mm}
	\label{tab:fid-scores}
	\tabcolsep=1mm
	\resizebox{0.64\linewidth}{!}{
    \begin{tabular}{l|l|l|c|c|c|l}
        \# & Training set & Test set(s) & Type & Mask input & Style input & FID \\
        \hline
        1 & COCO-train & COCO-val & 1-step \cite{park2019spade} & Ground truth & None & 22.64 \\
        2 & COCO-train & COCO-val & 1-step$\,^\dagger$ & Semantic seg. & None & 23.97 \\
        3 & COCO-train & COCO-val & 1-step$\,^\dagger$ & Sparse (ours) & None & 20.02 \\
        4 & VG+ (aug.) & COCO-val/VG-val & 1-step$\,^\dagger$ & Sparse (ours) & None & \textbf{18.93}/13.23 \\
        \hline
        5 & COCO-train & COCO-val & 1-step$\,^\dagger$ & Sparse (ours) & Text (6h, $L_{n-1}$) & 19.65 \\
        6 & COCO-train & COCO-val & 1-step$\,^\dagger$ & Sparse (ours) & Text (12h, $L_n$) & 20.63 \\
        7 & COCO-train & COCO-val & 2-step$\,^\dagger$ & Sparse (ours) & Text (12h, $L_n$) & 20.64 \\
        \hline
        8 & VG & COCO-val/VG-val & 1-step$\,^\dagger$ & Sparse (ours) & Attributes & 21.13/15.12 \\
        9 & VG & COCO-val/VG-val & 2-step$\,^\dagger$ & Sparse (ours) & Attributes & 20.83/14.88 \\
        
        \hline
    \end{tabular}
    }
    \resizebox{0.32\linewidth}{!}{
    \begin{tabular}{l|l|l|c}
         & Ref. & Experiment & FID ($\Delta$) \\
        \hline
        I & \#1 & COCO ``things'' only & 32.31 (+9.67) \\
        \hline
        II & \#6 & 12h, $L_{n}$, attr. in $D$ & 20.44 (-0.19) \\
        III & \#6 & 12h, $L_{n-1}$ & 19.77 (-0.86) \\
        IV & \#6 & 6h, $L_{n-1}$ & 19.65 (-0.98) \\
        \hline
        V & \#9 & No f.g. info in $S_{avg}$ & 25.16 (+4.33) \\
        VI & \#9 & Attr. randomization & 20.64 (-0.19) \\
        \hline
    \end{tabular}
    }
\end{table}

\begin{figure}[t]
    \centering
    \includegraphics[width=\textwidth]{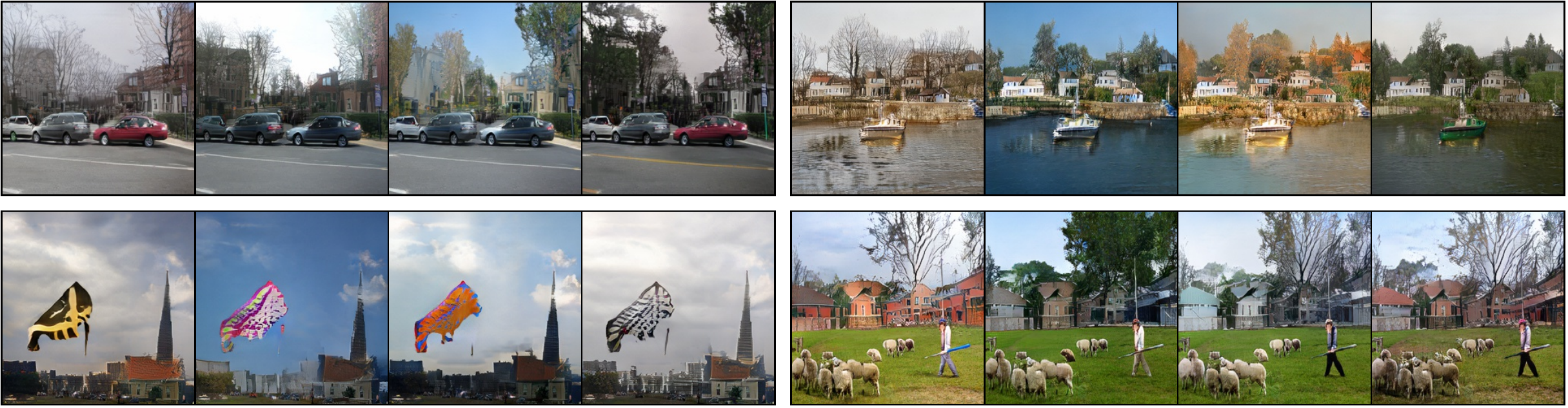}
	\caption{\small Random styles by sampling attributes from a per-class empirical distribution.}
	\label{fig:attribute-randomization}
\end{figure}

\paragraph{Style randomization.} Since we represent style explicitly, at inference we can randomize the style of an image by drawing attributes from a per-class empirical distribution. This is depicted in \autoref{fig:attribute-randomization}, and has the additional advantage of being interpretable and editable (attributes can be refined manually after sampling). The two-step decomposition also allows users to specify different sampling strategies for the background and foreground; more details in the Appendix \appref{sec:appendix-inference}{A.2}.

\paragraph{Ablation study.} While \autoref{tab:fid-scores} (left) already includes a partial ablation study where we vary input conditioning and some aspects of the attention module, in \autoref{tab:fid-scores} (right) we make this more explicit and include additional experiments.
First, we train a model on a sparsified COCO dataset by only keeping the ``things'' classes and discarding the ``stuff'' classes. This setting (I) performs significantly worse than \#1 (which uses all classes), motivating the use of a large class vocabulary.
Next, we ablate conditioning via text (baseline \#6, which adopts the default hyperparameters of BERT). In (II), we augment the discriminator with ground-truth attributes to provide a stronger supervision signal for the generator (we take the attributes from Visual Genome for the images that overlap between the two datasets). The improvement is marginal, suggesting that our model can learn visual-textual correlations without explicit supervision. In (III), we draw the token representations from the second-to-last layer instead of the last, and in (IV) we further reduce the number of attention heads from 12 to 6. Both III and IV result in an improvement of the FID, which justifies the hyperparameters chosen in \#5.
Finally, we switch to attribute conditioning (baseline \#9). In (IV), we remove foreground information at inference from the $S_{avg}$ block of the first generator $G_1$ (we feed the background mask twice in $S$ and $S_{avg}$). The FID degrades significantly, suggesting that $G_1$ effectively exploits foreground information to bias the result. In (V) we show that randomizing style at inference (previous paragraph) is not detrimental to the FID, but in fact seems to be slightly beneficial, probably due to the greater sample diversity.

\paragraph{Robustness and failure cases.} %
Input masks can sometimes be noisy due to spurious object detections on certain classes. Since these are also present at train time, weakly-supervised training leads to some degree of noise robustness, but sometimes the artifacts are visible in the generated images. We show some positive/negative examples in the Appendix Fig.\ \appref{fig:appendix-robustness}{14}. In principle, mask noise can be reduced by using a better object detector. We also observe that our setup tends to work better on outdoor scenes and sometimes struggles with fine geometric details in indoor scenes or photographs shot from a close range.

\section{Conclusion}
\label{sec:conclusion}

We introduced a weakly-supervised approach for the conditional generation of complex images. The generated scenes can be controlled through various manipulations on the sparse semantic maps, as well as through textual descriptions or attribute labels. Our method enables a high level of semantic/style control while benefiting from improved FID scores. From a qualitative point-of-view, we have demonstrated a wide variety of manipulations that can be applied to an image. Furthermore, our weakly supervised setup opens up opportunities for large-scale training on unlabeled datasets, as well as generation from hand-drawn sketches.

There are several ways one could pursue to further enrich the set of tools used to manipulate the generation process. For instance, the current version of our attention mechanism cannot differentiate between instances belonging to the same class and does not have direct access to positional information. 
While incorporating such information is beyond the scope of this work, we suggest that this can be achieved by appending a \emph{positional embedding} to the attention queries. In the NLP literature, the latter is often learned according to the position of the word in the sentence \cite{vaswani2017transformer,devlin2018bert}, but images are 2D and therefore do not possess such a natural order. Additionally, this would require captions that are more descriptive than the ones in COCO, which typically focus on actions instead of style. Finally, in order to augment the quality of sparse maps, we would like to train the object detector on a higher-quality, large-vocabulary dataset \cite{gupta2019lvis}.

\paragraph{Acknowledgments.} This work was partly supported by the Swiss National Science Foundation (SNF) and Research Foundation Flanders (FWO), grant \#176004. We thank Graham Spinks and Sien Moens for helpful discussions.

\clearpage
\bibliographystyle{splncs04}
\bibliography{egbib}

\ifdefined\preprint
\newpage
\clearpage
\newpage
\appendix

\section{Supplementary material}
\subsection{Detailed architecture}
\label{sec:appendix-architecture}
\begin{wrapfigure}{r}{0.4\textwidth}
    \centering
    \vspace{-25mm}
    \includegraphics[width=0.6\linewidth]{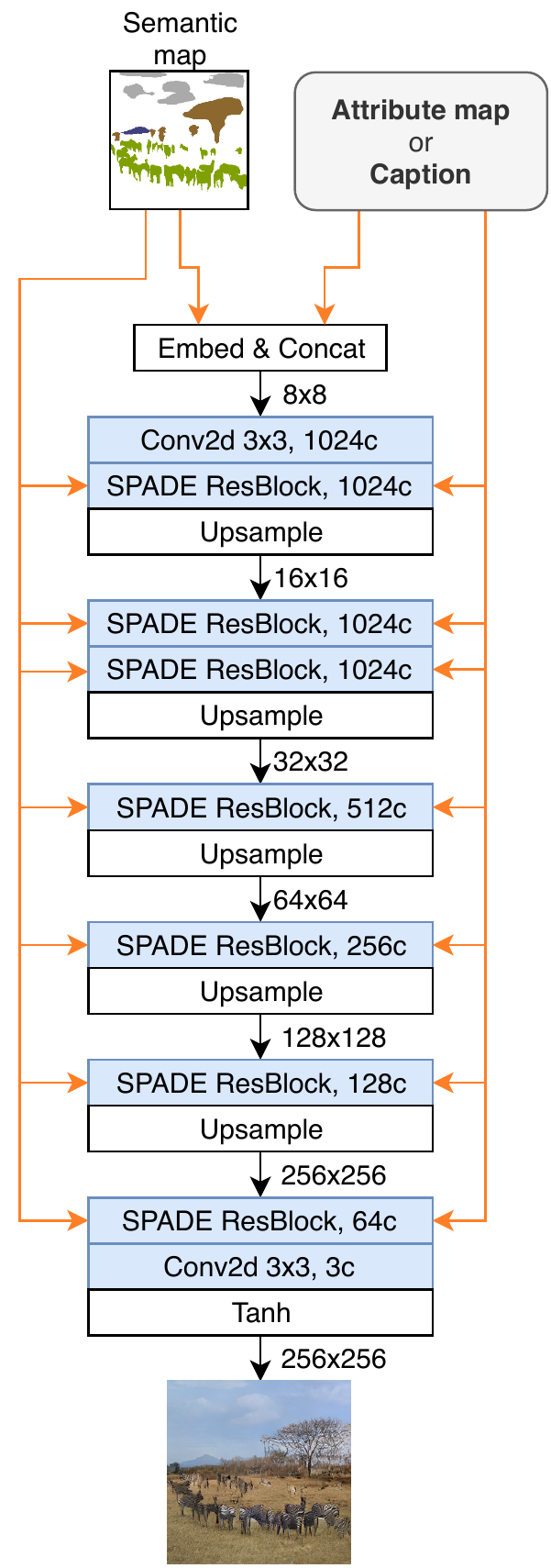}\\
    \vspace{5mm}
    \includegraphics[width=0.7\linewidth]{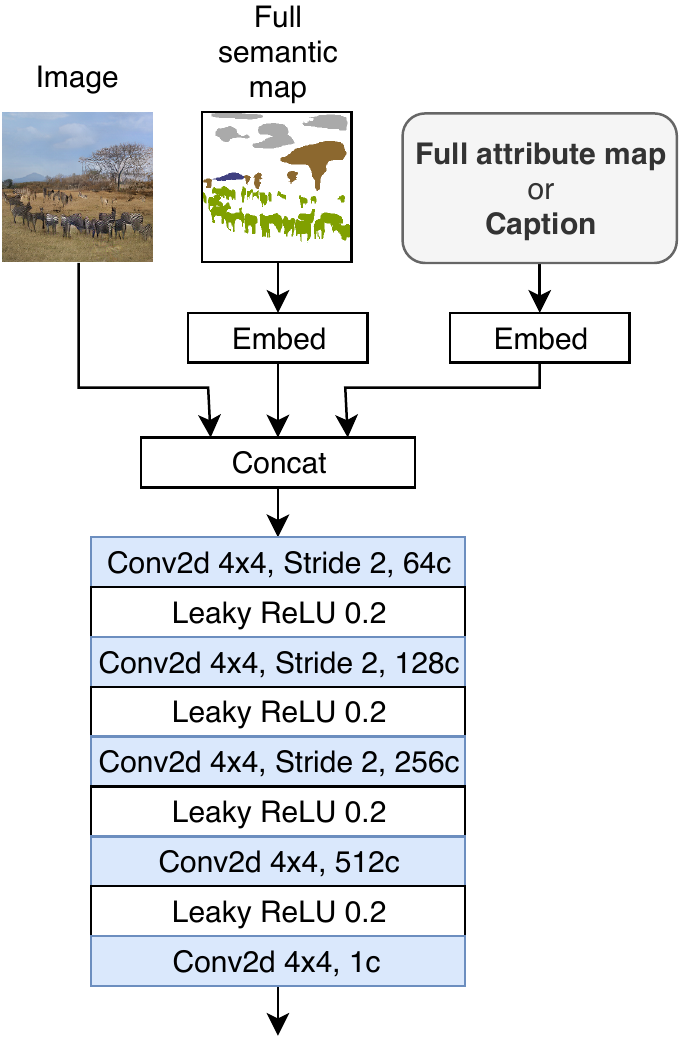}
    \caption{\small \textbf{Top:} one-step generator using the SPADE backbone. ``1024c'' stands for ``1024 output channels''. The number on the right of an arrow specifies the feature map resolution at that level. Orange arrows indicate that the input information is fed to $S$ blocks. \textbf{Bottom:} discriminator (used in all architectures).}
    \label{fig:appendix-1step}
\end{wrapfigure}

In this section, we provide additional implementation details about our architecture in order to consolidate the already-presented \autoref{fig:architectures} (overview of the generators) and \autoref{fig:blocks} (conditioning blocks).

\paragraph{One-step generator.}
In \autoref{sec:architecture} we mention that we use \cite{park2019spade} as the backbone for the one-step model, and that we insert conditioning information in the normalization blocks as well as in the very first layer of the generator. In \autoref{fig:appendix-1step} (top) we show the detailed architecture of this model. The implementation of an individual ``SPADE ResBlock'' is specified in \cite{park2019spade}, but for reference we mention that each residual block consists of two normalization blocks wrapped by a skip-connection. If the number of input and output channels does not match, the skip-connection is learned, i.e. a third normalization block is learned. In the models conditioned on captions, we never attach attention inputs to skip-connections (to avoid potential instabilities). Each normalization block learns its own set of weights, and in our case they correspond to the $S$ or $S_{avg}$ blocks specified in \autoref{fig:blocks}.

\paragraph{Two-step generator.}
The architecture of the two-step generator is depicted in \autoref{fig:appendix-generator-2step}, and differs significantly from the aforementioned implementation. The background generator $G_1$ is a simplified version of the one-step generator with fewer residual blocks. The foreground generator $G_2$ implements a bottleneck architecture that takes as input the generated background image and compresses it through a series of \emph{unconditional} residual blocks. The low-resolution feature-map is then expanded again through a series of \emph{conditional} blocks. Interestingly, for foreground manipulations it is possible to preprocess the feature maps up to the last \emph{unconditional} downsampling block in $G_2$ ($8 \times 8$ resolution) and greatly speed up regeneration. 

\begin{wrapfigure}{r}{0.55\textwidth}
    \centering
    \includegraphics[width=\linewidth]{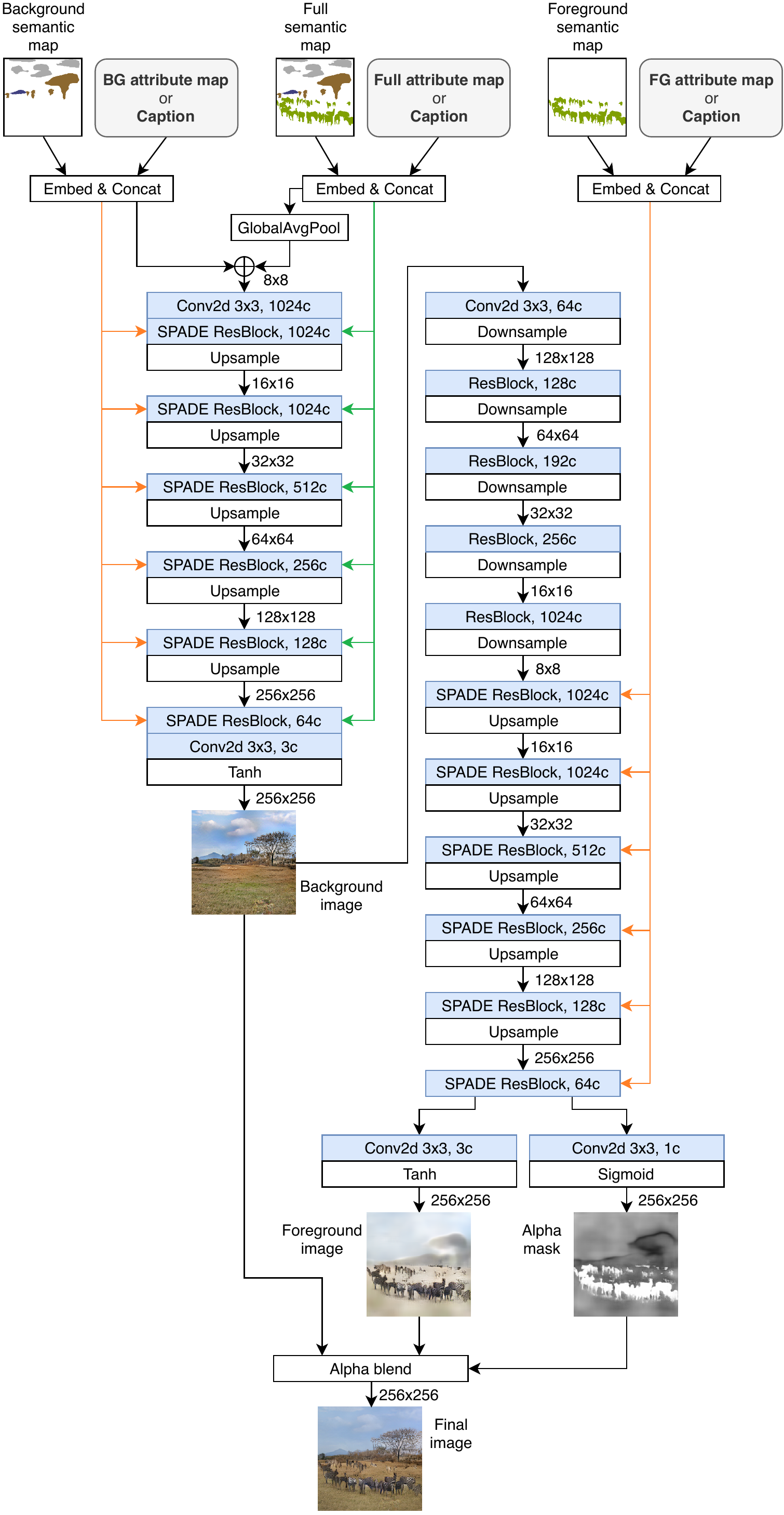}
	\caption{\small Two-step generator. The left side of the figure depicts $G_1$ (background generator), while the right side depicts $G_2$ (foreground generator). Orange arrows indicate that the input information is fed to $S$ blocks, whereas green arrows denote inputs to $S_{avg}$ blocks.}
	\label{fig:appendix-generator-2step}
	\vspace{-3mm}
\end{wrapfigure}

\paragraph{Discriminator.}
We use the multi-scale discriminator from \cite{wang2018pix2pixhd,park2019spade} and change its input layer to add information about attributes or captions. The architecture is shown in \autoref{fig:appendix-1step} (bottom). As usual with multi-scale discriminators, we train two instances: one which takes as input an image at full resolution, and one which takes as input a downsampled version (by a factor of two). They learn different sets of embeddings and different sets of attention heads if the style is conditioned on a sentence.

\paragraph{Model complexity.}
\autoref{tab:parameters} presents the number of parameters for all variants of our approach. The SPADE baseline trained on the 182 COCO-Stuff classes requires 97.5M parameters. Our 1-step baseline trained without style information (neither attributes nor captions) on our set of 280 classes requires a slightly lower number of parameters (94.2M) thanks to the pixel-wise class embeddings, even though the number of classes is larger. 
In the version with attributes, the added cost (+2.3M parameters) is only due to the learned attribute embeddings (256 64d embeddings per normalization block). In the version with captions, the custom attention modules add 12.5M parameters (for 6 heads) or 23.3M parameters (for 12 heads). The number of parameters can be easily tuned by varying the number of attention heads.
We conduct a similar analysis on the two-step model. In this case, the background generator is slightly more powerful than the foreground generator.

\begin{table}[htb]
    \centering
    \caption{\small Number of parameters for different variations of our approach. For the two-step models we specify the numbers for both generators (respectively $G_1$ and $G_2$). ``6h'' denotes ``6 attention heads''.}
    \label{tab:parameters}
    \vspace{2mm}
	\tabcolsep=1mm
	\resizebox{0.45\linewidth}{!}{
    \begin{tabular}{c|c|c}
        Approach & Style input & \# params \\
        \hline
        Baseline \cite{park2019spade} & None & 97.5M \\
        \hline
        1-step & None & 94.2M \\
        1-step & Attributes & 96.5M \\
        1-step & Text (6h) & 106.7M \\
        1-step & Text (12h) & 117.5M \\
        \hline
        2-step & None & 74.5M + 50.6M\\
        2-step & Attributes & 78.3M + 51.9M \\
        2-step & Text (12h) & 90.7M + 65.8M \\
        \hline
    \end{tabular}
    }
\end{table}

\newpage
\paragraph{Sparse map generation and manipulation.}
In this paragraph we provide further details in addition to those presented in \autoref{sec:implementation-details}. Specifically, we describe how we construct and maintain the data structure that enables instance manipulation and rasterization into a sparse semantic map. Since a scene may consist of objects that partially overlap, the order in which they are drawn on the semantic map matters, e.g. given a \emph{car} and its \emph{headlight}, we want to render the headlight semantic mask on top of the car and not the opposite. Therefore, we sort all instances by mask area and draw them from the largest to the smallest. Additionally, we construct a scene graph to facilitate manipulation: if 70\% of the area of an instance is contained within another instance, it becomes a child of the latter. With regard to the previous example, moving the car would also move the headlights attached to it. %
Finally, in our experiments on Visual Genome, we link attributes to an instance if the IoU between the ground-truth region and the detected bounding box is greater than 0.5.

\paragraph{Training details and hyperparameters.} In all experiments, we train on 8 Pascal GPUs for 100 epochs using Adam (learning rate: 1e-4 for $G$, 4e-4 for $D$, one $G$ update per $D$ update), and start decaying the learning rate to 0 after the 50th epoch in a linear fashion. We use a batch size of 32 for the \emph{one-step} model and 24 for the \emph{two-step} model (the largest we can fit into memory), with synchronized batch normalization. Training takes one week for the one-step model and two weeks for the two-step model. For the alpha blending loss term, we start from a factor of 10, and decay it exponentially with $\alpha = 0.9997$ per weight update, down to 0.01. For the experiments with captions, since COCO comprises five captions per image, we randomly select one caption at training time. In the evaluation phase, we concatenate the representations of all captions since our attention model can easily decide which ones to attend to.

\subsection{Additional inference details}
\label{sec:appendix-inference}

\begin{figure*}[!b]
    \centering
    \includegraphics[width=\textwidth]{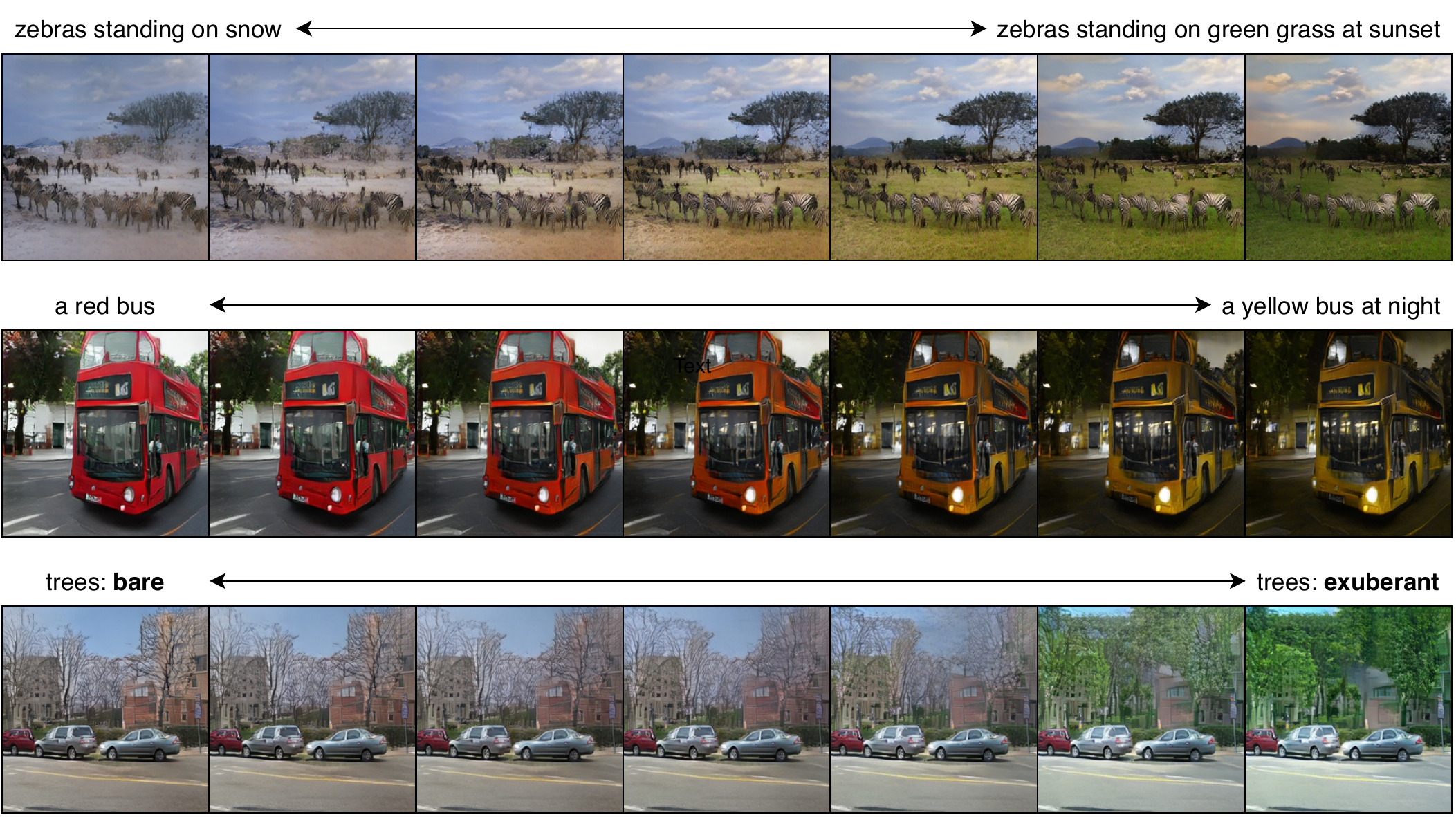}
	\caption{\small Interpolating style between two sentences (top two rows) and two attributes (bottom row). The smooth transitions across multiple factors of variation (e.g. color and time of the day) suggest that our latent space is structured and does not require regularization. For instance, in the middle row, the bus color traverses the region of \emph{orange} while interpolating between red and yellow, even though it is not explicitly instructed to do so. Additionally, the headlights of the bus become increasingly brighter.}
	\label{fig:interpolation}
\end{figure*}

\paragraph{Randomizing style.} In \autoref{sec:results} we mention that we can randomize the style of an image by sampling attributes from a per-class empirical distribution. More precisely, we estimate a discrete probability distribution of the attributes assigned to each class of the dataset. This includes the empty set (no attribute for a given instance) as well as compound attributes (e.g. \emph{blue and red} is different than \emph{blue} or \emph{red}). At inference, for each instance, we sample an element from the distribution of the class to which the instance belongs. The two-step decomposition also allows us to specify different strategies for the background and foreground. In the examples in \autoref{fig:attribute-randomization}, all background instances of a given class take the same attributes as input (e.g. all trees are leafless), which results in scenes with coherent styles. Conversely, foreground instances are still fully randomized (it would not be realistic to see cars all of the same color, for example). Within an individual instance, the style of its children is uniform, e.g. the same attributes are assigned to all wheels of a car, but of course wheel styles can be different across different cars.

\paragraph{Interpolating style.} Our approach allows for smooth interpolation of attributes and text. While attention models usually preclude interpolation (whereas models based on fixed-length sentence embeddings such as \cite{zhang2017stackgan} easily allow it), our \emph{sentence-semantic} attention mechanism enables interpolation over the \emph{contextualized class embeddings}, i.e. over the pooled attention values. For all cases (masks, attributes, text), we respectively interpolate between class embeddings, attribute embeddings, and contextualized class embeddings using spherical interpolation (\emph{slerp}), which traverses regions with a higher probability mass \cite{kilcher2018semantic}. Unlike \cite{zhang2017stackgan}, we found it unnecessary to enforce a prior on the embeddings via a KL divergence term in the loss. We show some examples of interpolation in \autoref{fig:interpolation} as well as in the supplementary video (\autoref{sec:appendix-video}).

\paragraph{Generating one object at a time.} %
To ensure that foreground objects do not affect each other in the two-step model, it may be interesting to generate them one-by-one. In our experiments we generate all foreground objects at once by running a single instance of $G_2$, motivated by the much lower computational cost and the observation that foreground objects are usually well-separated. Nonetheless, our framework is flexible enough to support one-by-one generation of objects. In this regard, $G_2$ can be run independently for each object, and the output images and masks can be combined into a single, final image. Denoting the background image as $\mathbf{x}_{\text{bg}}$, the foreground images as $\mathbf{x}^{[i]}_{\text{fg}}$ ($i \in \{1 \ldots N\}$), and the corresponding \emph{unscaled} (i.e. before the activation function) transparency masks as $\boldsymbol{\alpha'}^{[i]}_{\text{fg}}$, we can generalize \autoref{eq:alphablend} as follows:

\begin{align}
    \mathbf{w}^{[i]}_{\text{fg}} &= \text{softmax}_{\,i} \left(\boldsymbol{\alpha'}^{[i]}_{\text{fg}}\right)\\
    \mathbf{x}_\text{fg} &= \sum_i \mathbf{x}^{[i]}_{\text{fg}} \odot \mathbf{w}^{[i]}_{\text{fg}}\\
    \boldsymbol{\alpha}_\text{fg} &= \sum_i \text{sigmoid}\left(\boldsymbol{\alpha'}^{[i]}_{\text{fg}}\right) \odot \mathbf{w}^{[i]}_{\text{fg}}\\
    \textbf{x}_{\text{final}} &= \textbf{x}_{\text{bg}} \cdot (1 - \boldsymbol{\alpha}_{\text{fg}}) + \textbf{x}_{\text{fg}} \cdot \boldsymbol{\alpha}_{\text{fg}}
\end{align}
The second line combines foreground images into a single image through an object-wise weighted average. The same is repeated for the transparency channel (third line). Finally, the alpha blending is performed as in \autoref{eq:alphablend}. This formulation is differentiable and can be used for training the model, although the memory requirement may be excessive in high-resolution settings.

\subsection{FID evaluation}
\label{sec:appendix-fid}

The FID metric is very sensitive to aspects such as image resolution, number of images (where a low number results in underestimated FID scores), and the weights of the pretrained Inception network. To be consistent with \cite{park2019spade}, we try to follow their methodology as closely as possible. We resize the ground-truth images to the same resolution as the generated ones ($256 \times 256$), and we keep the two sets aligned, i.e. one generated image per test image. We use the weights of the pretrained InceptionV3 network provided by PyTorch. To make the results in \autoref{tab:fid-scores} comparable, we retrained the baseline from \cite{park2019spade} and evaluated the results using our methodology.

\begin{figure}[!t]
    \centering
    \includegraphics[width=\textwidth]{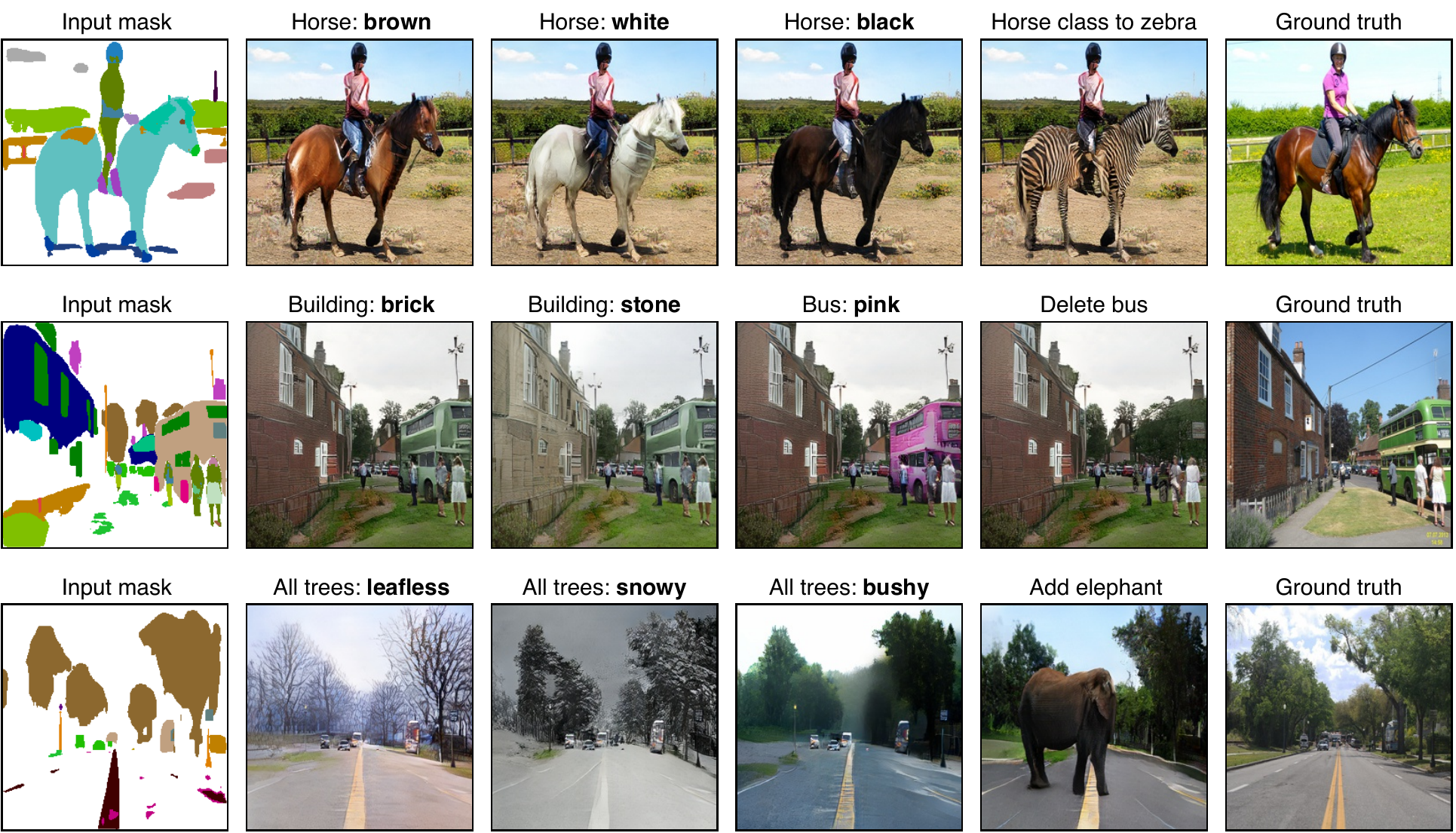}
	\caption{\small Examples of semantic and attribute manipulations (Visual Genome dataset). The images are generated by our two-step model. In the first row, the background is frozen to encourage locality.}
	\label{fig:appendix-demos-attributes}
\end{figure}

\begin{figure}[!t]
    \centering
    \includegraphics[width=\textwidth]{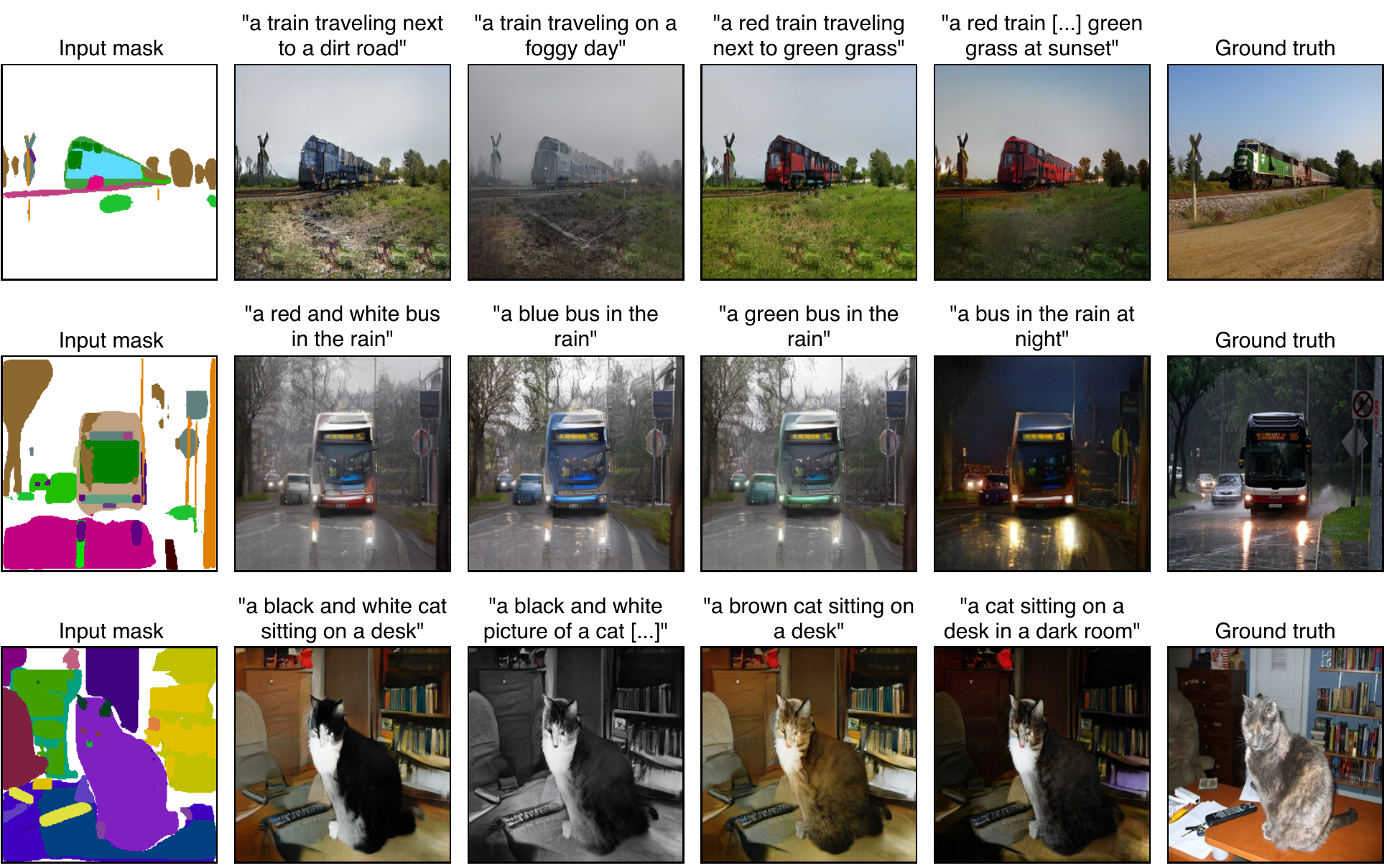}
	\caption{\small Further examples of style manipulation using text (COCO validation set). It is possible to control the style of individual instances (albeit in a less targeted fashion than attributes) as well as the global style of the image.}
	\label{fig:appendix-demos-text}
\end{figure}

\subsection{Additional results}
\label{sec:appendix-results}
\paragraph{Semantic and style manipulation.}
\autoref{fig:appendix-demos-attributes} and \autoref{fig:appendix-demos-text} show examples of semantic manipulation and style manipulation (either using attributes or text). The last row of \autoref{fig:appendix-demos-text} suggests that our attention mechanism can correctly exploit the contextualized token embeddings produced by BERT. For instance, the caption ``a black and white cat'' affects only the cat, while ``a black and white picture of a cat'' affects the entire scene by generating a black-and-white image.\looseness=-1

\paragraph{Two-step model.}
\autoref{fig:appendix-demos-2step} shows additional demos generated by our two-step model on the Visual Genome validation set. In particular, we highlight the decomposition of the background and foreground, and the inputs taken by $G_1$ and $G_2$. Since $G_2$ outputs a soft transparency channel for the alpha blending, it can slightly violate the constraints imposed by the \emph{foreground mask}. This allows it to draw reflections and shadows underneath foreground objects. Furthermore, as we mention in \autoref{sec:framework}, the motivation behind the two-step generator is that it facilitates local changes. In \autoref{fig:appendix-local-global} we qualitatively compare one-step and two-step generation when manipulations are carried out on the input conditioning information (mask and style). We show that, in the two-step model, local manipulations do not result in global changes of the output. To further enhance locality, the background can be frozen when manipulating the foreground.

\begin{table}[h]
    \centering
    \caption{\small Comparison to layout-based methods. The metric is the FID score \cite{heusel2017ttur}; lower is better. ``GT BBox'' stands for ``ground-truth bounding-box'', whereas our approach uses the sparse masks inferred from an object detector as usual. }
    \label{tab:extra-comparison}
    \vspace{2mm}
	\tabcolsep=1mm
	\resizebox{0.6\linewidth}{!}{
    \begin{tabular}{l|c|c|c|c}
        Approach & Input & Training set & Test set & FID \\
        \hline
        Sg2im \cite{johnson2018imagescenegraphs} & GT BBox layout & COCO-train & COCO-val & 67.96 \\
        Layout2im \cite{zhao2019imagelayout} & GT BBox layout & COCO-train & COCO-val & 38.14 \\
        LostGAN \cite{sun2019image} & GT BBox layout & COCO-train & COCO-val & 34.31 \\
        \hline
        Ours (\#3) & Sparse mask & COCO-train & COCO-val & 18.57 \\
        Ours (\#5) & Sparse mask & VG+ (aug.) & COCO-val & \textbf{17.98}
    \end{tabular}
    }
\end{table}

\paragraph{Comparison with layout-based methods.} While in \autoref{sec:results} we compare our approach to \cite{park2019spade} under uniform settings, it is also interesting to see how our sparse mask setting compares to approaches that generate images from bounding-box layouts (which are also sparse by nature) \cite{hong2018inferring,zhao2019imagelayout,sun2019image}. While these methods address a harder task (bounding boxes provide less information than segmentation masks), their applicability has only been demonstrated in low-resolution settings (typically $64 \times 64$), which makes them not directly comparable to our higher-resolution setting. To our knowledge, no bounding-box approach can currently generate high-resolution images that have the same visual quality and geometric coherence as mask-based approaches. Nonetheless, for completeness, in \autoref{tab:extra-comparison} we compare our sparse mask approach to these layout-based methods. We use the models trained on COCO or VG+ with no style input (rows \#3 and \#4 in \autoref{tab:fid-scores}, left), and downscale our images to $64 \times 64$ before computing the FID score.

\begin{figure*}[!b]
    \centering
    \includegraphics[width=\textwidth]{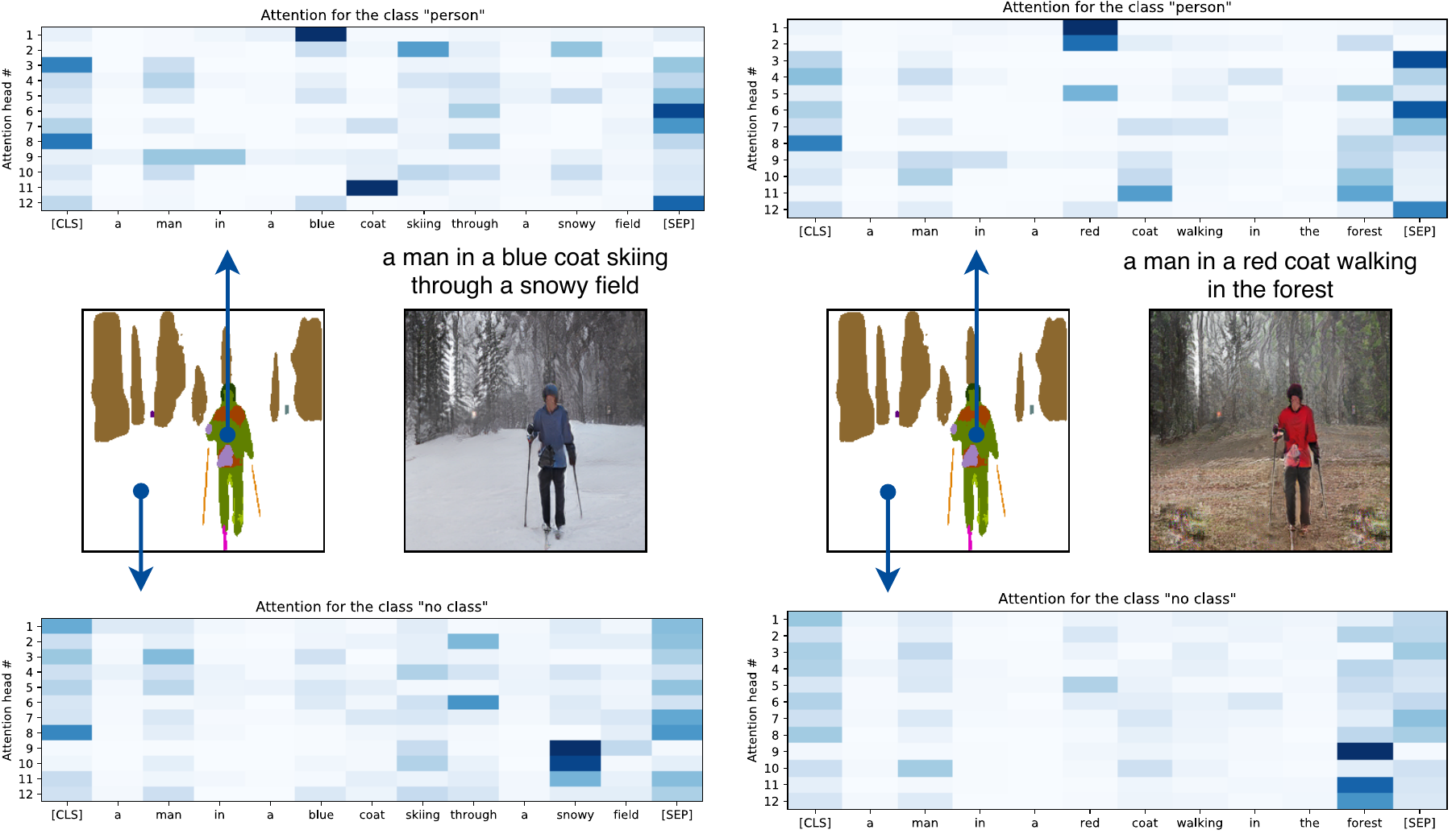}
	\caption{\small Visualization of the attention mechanism in the discriminator for two images generated from the same semantic map, but different captions. An attention map is produced for each class in the semantic map, and each of these consists of 6 or 12 independent attention heads (12 here). In this illustration we only show those corresponding to \emph{person} and \emph{no class} (i.e. blank space) for clarity. \texttt{[CLS]} and \texttt{[SEP]} are special delimiters indicating respectively the start and end of a sentence. A head paying attention to these can be interpreted as not being triggered by the sentence. In the attention maps, a darker color indicates a higher weight.}
	\label{fig:appendix-attention}
\end{figure*}

\paragraph{Qualitative comparison of input masks.} In \autoref{fig:appendix-mask-comparison}, we show qualitative results for different input masks, both in fully supervised and weakly supervised settings. Additionally, in the figure we show qualitative results for the \emph{sparsified} COCO model (ablation I in \autoref{tab:fid-scores}, right), where we keep only the ``thing'' classes of COCO. While the outputs produced by the semantic segmentation maps are satisfactory, it is not clear how to manipulate them as they present banding artifacts and jagged edges.

\subsection{Attention visualization}
\label{sec:appendix-attention}
The behavior underlying our attention model can be easily visualized. Our formulation (\emph{sentence-semantic} attention) is particularly suited for visualization tasks because it is tied to the semantic map, and not to feature maps in inner convolutional layers. Therefore, for each class in the semantic map (e.g. \emph{person}, \emph{tree}, empty space), we can observe how the sentence conditions that particular class.
Considering that the attention modules have multiple entry points in the generator (one for each normalization block), it is easier to carry out this analysis in the discriminator, where there are only two entry points (in the input layer of each discriminator, since we adopt a multi-scale discriminator). We select the first discriminator for illustration purposes, and show the resulting attention maps in \autoref{fig:appendix-attention}. The figure shows what parts of the sentence the discriminator is attending to in order to discriminate whether the caption is suitable for the input image.

\begin{figure}[tb]
    \centering
    \includegraphics[width=\textwidth]{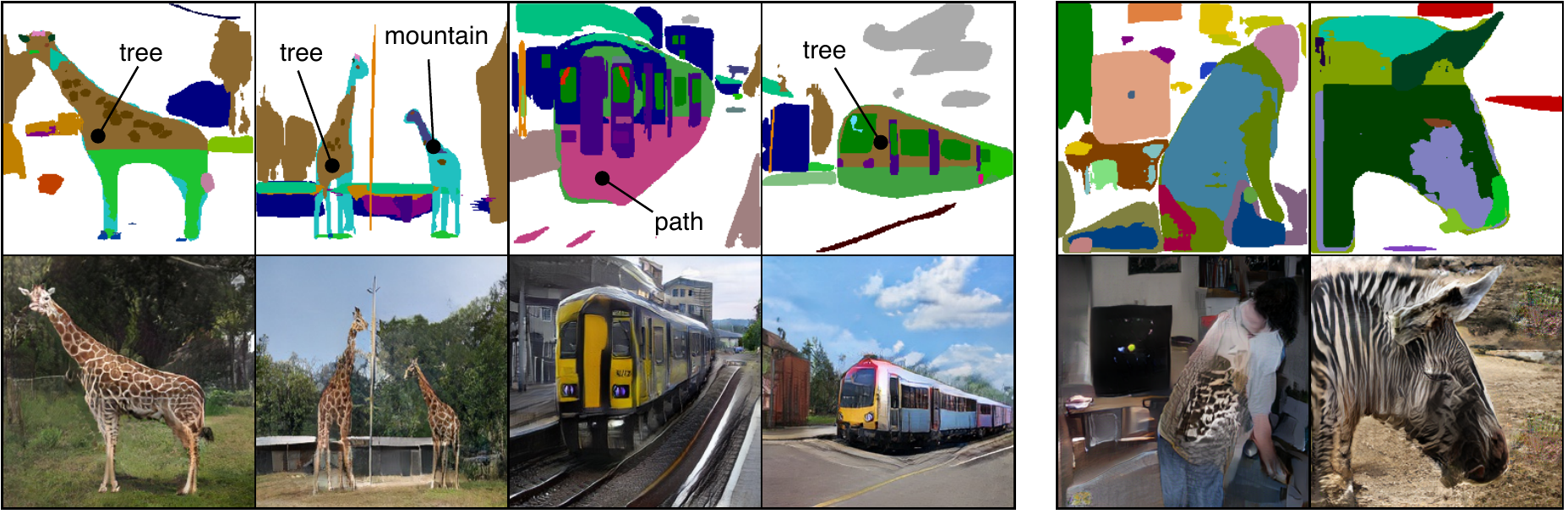}
	\caption{\small \textbf{Left:} in many cases, weakly-supervised training leads to input noise robustness, i.e. artifacts in the input mask are not visible in the generated images. \textbf{Right:} some failure cases where the artifacts are visible in the output images.}
	\label{fig:appendix-robustness}
	\vspace{-3mm}
\end{figure}

\subsection{Negative results}
\label{sec:appendix-negative-results}
In this section, we discuss some of the unsuccessful ideas that we explored before reaching our current formulation.

\paragraph{Two-step model.} Before successfully achieving two-step generation with sparse masks, we tried to implement the same idea using dense COCO segmentation maps. In the areas corresponding to foreground objects, $G_1$ (the background generator) would always render visible gaps. We tried to regularize the model using \emph{partial convolutions} (a recently-proposed approach for infilling), but this did not have the desired effect. We also experimented with an attention mechanism where foreground areas were masked in $G_1$. While this was partly successful in filling the gaps, the model was very difficult to train and the final visual quality was considerably lower.

\paragraph{Discriminator architecture.} We explored various ways of injecting conditional information in the discriminator. While SPADE uses input concatenation, recent GANs conditioned on classes \cite{zhang2018sagan,brock2018biggan} use \emph{projection discrimination} \cite{miyato2018cgans}. This idea led to marginally better FID scores, but we observed that the contour of generated objects would stick too close to the input mask, essentially resulting in a ``polygonal'' appearance. On the other hand, input concatenation allows the model to slightly deviate from the input mask, possibly resulting in a greater robustness to mask noise.

\paragraph{Hyperparameters.} We tried to vary the design of SPADE blocks, e.g. by stacking more layers or using dilated convolutions. These ideas had a detrimental effect on the final result and we decided not to pursue them further.

\subsection{Demo video}
\label{sec:appendix-video}
The video at \url{https://github.com/dariopavllo/style-semantics} illustrates examples of interactive manipulations. Among other things, we show how images can be generated from sketches as the user draws the masks, extra results from the two-step model (including comparisons with the one-step model with regard to local changes), and interpolations in the latent space of text and attributes.

\begin{figure*}[tb]
    \centering
    \includegraphics[width=\textwidth]{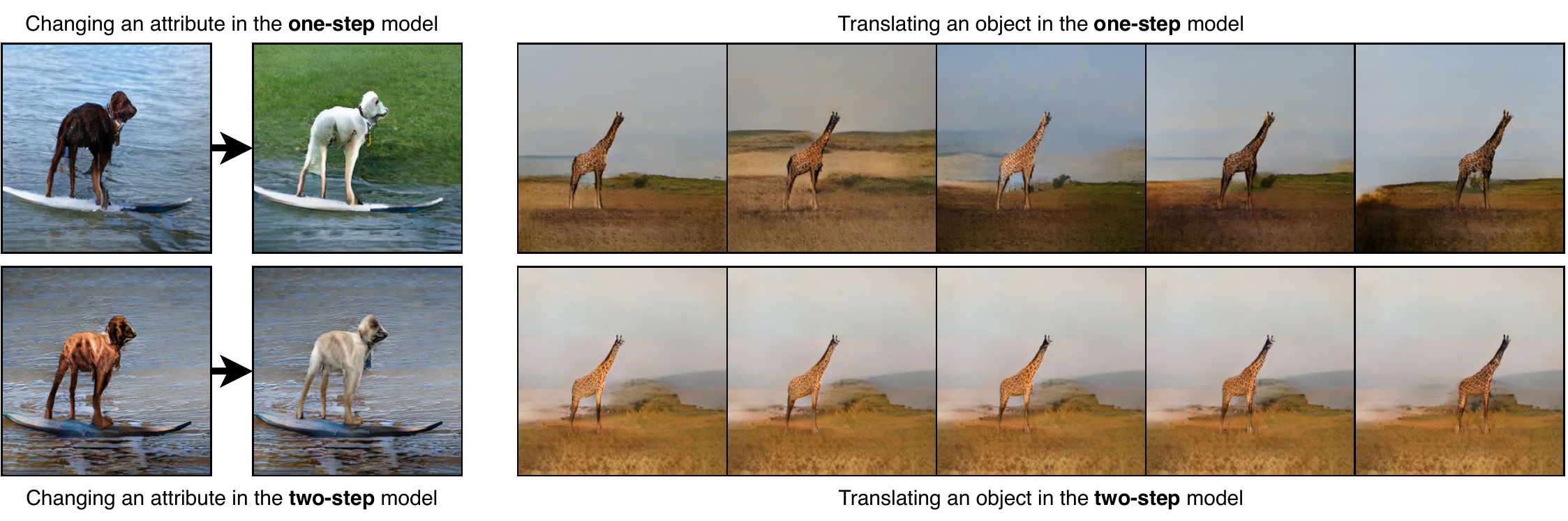}
	\caption{\small In a single-generator model, local changes (e.g. changing the color of the dog to white) affect the scene globally due to learned correlations. The same can be observed when moving an object (e.g. left to right), as the representation space is discontinuous. In the two-step model, we can locally manipulate the background and foreground.}
	\label{fig:appendix-local-global}
\end{figure*}
\begin{figure*}[tb]
    \centering
    \includegraphics[width=\textwidth]{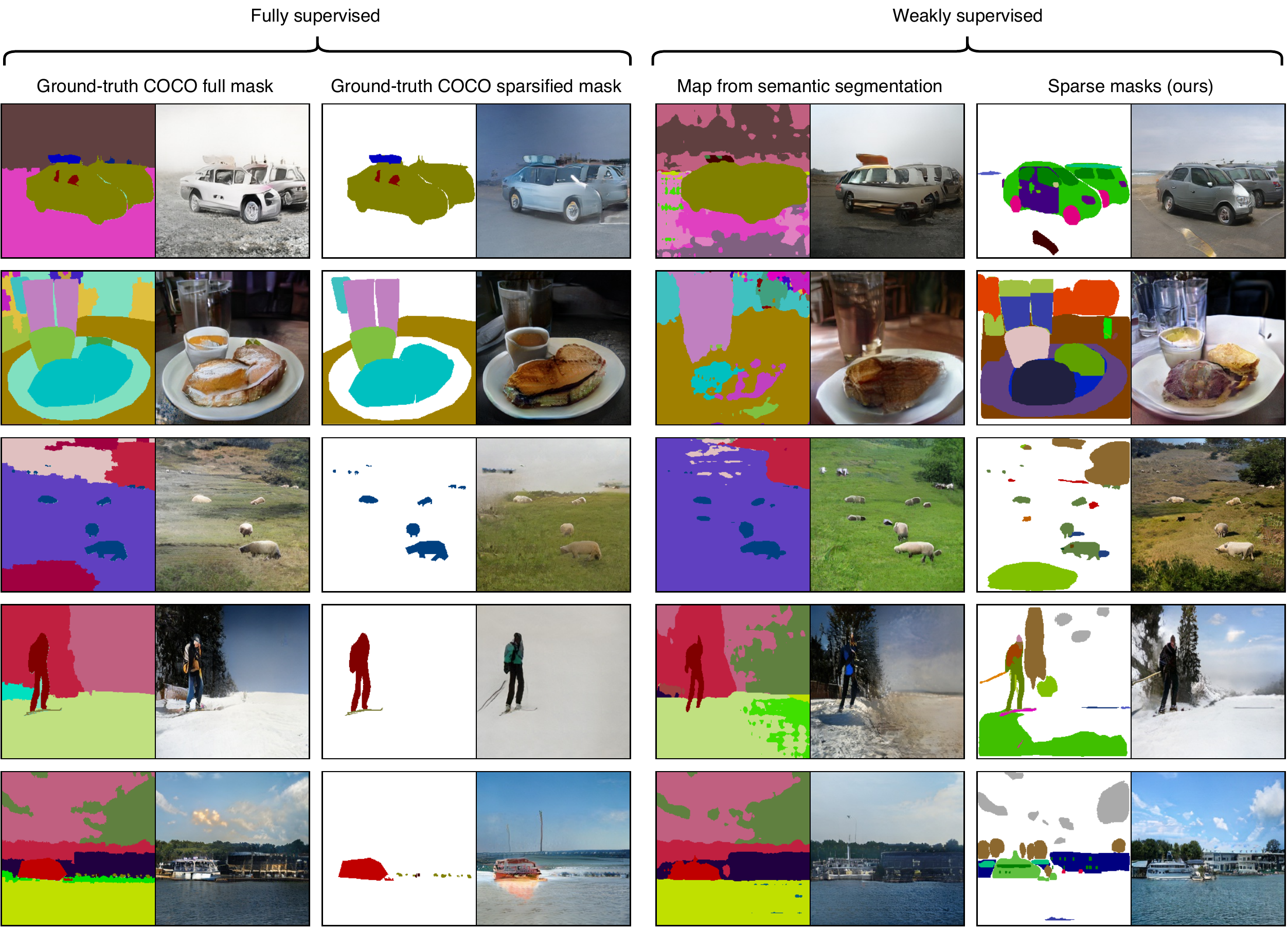}
	\caption{\small Input masks for different approaches, and corresponding generated images. Our sparse masks do not present the typical artifacts of semantic segmentation outputs and are much easier to sketch or manipulate than dense maps.}
	\label{fig:appendix-mask-comparison}
\end{figure*}

\begin{figure*}[tb]
    \centering
    \includegraphics[width=\textwidth]{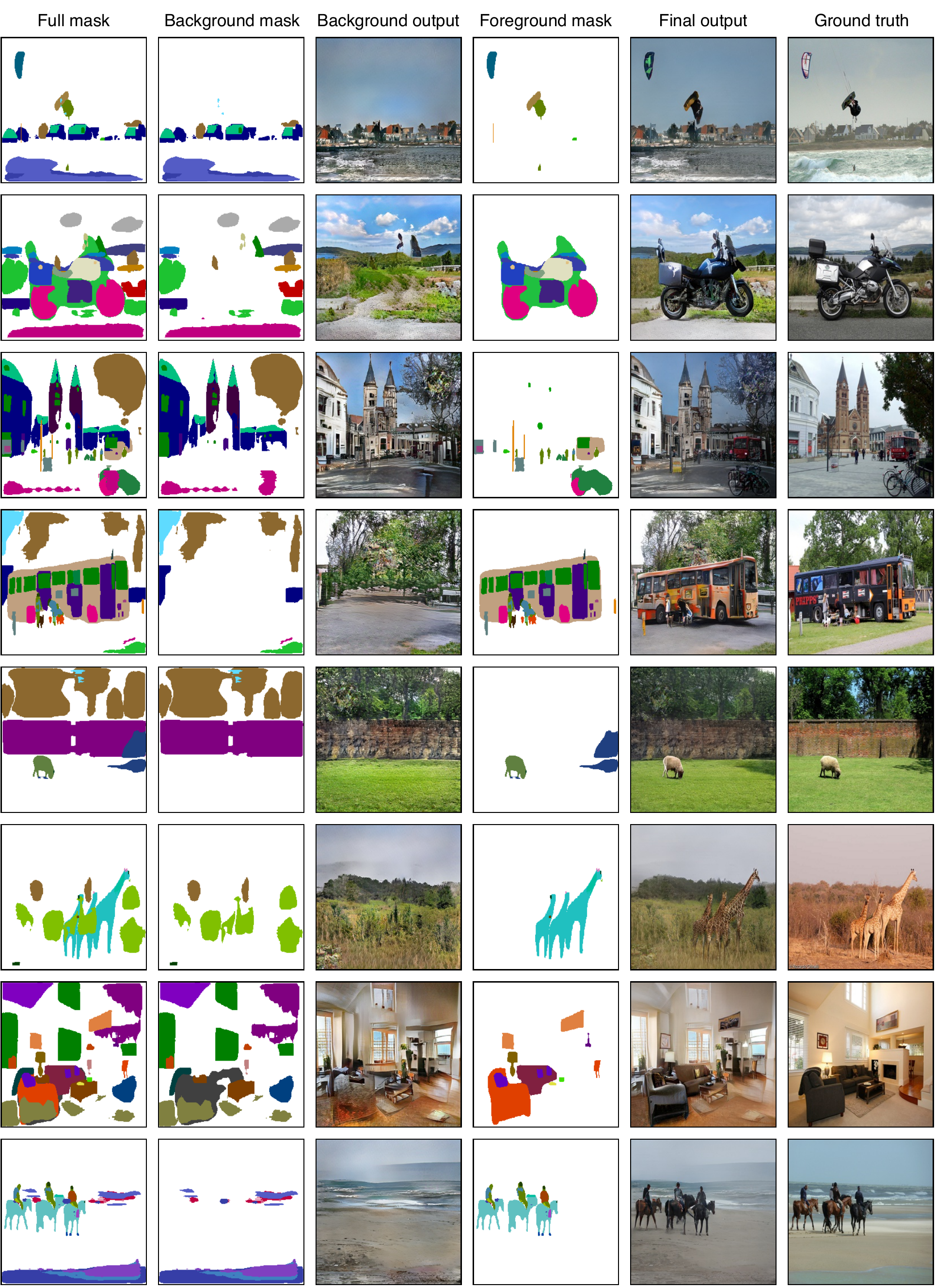}
	\caption{\small Demos generated by our two-step model. In addition to the full input mask, we show its decomposition into \emph{background mask} and \emph{foreground mask} (taken as input in $S$ blocks respectively by $G_1$ and $G_2$). Note that $G_1$ also takes as input the full mask in $S_{avg}$ blocks.}
	\label{fig:appendix-demos-2step}
\end{figure*}
\fi

\end{document}